\def\ARXIVCOMBINED{1}
\documentclass[runningheads]{llncs}

\usepackage{eccv}

\usepackage{eccvabbrv}

\usepackage{graphicx}
\usepackage{booktabs}
\usepackage{multirow}
\usepackage{amsmath}
\usepackage{amssymb}
\usepackage{docmute}
\usepackage{import}

\usepackage[accsupp]{axessibility}  %

\usepackage{hyperref}

\usepackage{orcidlink}

\usepackage{xcolor}
\definecolor{projectblue}{RGB}{55, 125, 188}
\hypersetup{
    colorlinks=true,
    urlcolor=projectblue,
    linkcolor=black,
    citecolor=black
}
\newcommand{\todo}[1]{\textcolor{red}{[TODO: #1]}}

\usepackage{bbm}

\begin{document}

\title{MAVIN: Multi-Shot Audio-Visual Generation with Customized Narrative Control} 

\titlerunning{MAVIN: Multi-Shot Audio-Visual Generation}

\author{
Kaiqi Liu\inst{1,2}$^{*}$ \and
Yunyao Mao\inst{2}$^{*}$ \and
Ziqi Cai\inst{1} \and
Zheng Geng\inst{3} \and
Jing Wang\inst{4} \and\\[-0.1em]
Qiulin Wang\inst{2} \and
Xintao Wang\inst{2} \and
Pengfei Wan\inst{2} \and
Kun Gai\inst{2} \and\\[-0.1em]
Shuchen Weng\inst{1}$^{\dagger}$ \and
Boxin Shi\inst{1}$^{\dagger}$
}

\authorrunning{K.~Liu et al.}

\institute{
Peking University \and
Kling Team, Kuaishou Technology \and
Institute of Automation, Chinese Academy of Sciences \and
Sun Yat-sen University
\\
\email{liukq04@gmail.com, \{shuchenweng, shiboxin\}@pku.edu.cn}
}
\maketitle

{\centering\small
Project page: \url{https://liukqchoco.github.io/MAVIN/}\par
}

\begingroup
\renewcommand{\thefootnote}{}
\footnotetext{$^{*}$ Equal contribution. $^{\dagger}$ Corresponding authors.}
\addtocounter{footnote}{-1}
\endgroup

\begin{abstract}
While recent generative models produce high-fidelity videos, they struggle with the complex narrative control required for coherent multi-shot audio-visual generation. 
Existing methods suffer from temporal misalignment, limited controllability, and incomplete scripting. 
In this paper, we propose MAVIN, the first framework for multi-shot audio-visual generation with customized narrative control. To resolve temporal misalignment, we propose boundary-aware attention, which leverages hierarchical captions and boundary-aware token routing to render audio-visual elements within their respective temporal boundaries. 
To improve the controllability for multi-subject scenarios, we propose ID-aware propagation, utilizing identity embeddings and an identity-aware mask to bind specific identities to consistent visual appearances and vocal timbres. 
To provide comprehensive audio-visual narratives, we present a multi-agent scripting pipeline to transform free-form user inputs into hierarchical captions. 
Furthermore, we construct MAVINSet, a multi-shot audio-visual dataset for robust training and evaluation. Extensive experiments demonstrate that MAVIN achieves state-of-the-art performance, opening up a new avenue for integrating generative models into professional filmmaking workflows.

\keywords{Audio-Visual Generation \and Video Diffusion Models}
\end{abstract}

\section{Introduction}
\label{sec:intro}

Recent video generation models~\cite{wan2025wan,hong2022cogvideo,gao2025seedance,yang2024cogvideox, wu2025hunyuanvideo} have achieved remarkable progress in generating high-fidelity videos. With data engineering and architectural adaptation, they are widely explored in specialized scenarios, enabling precise camera movement~\cite{Wang2024MotionCtrl,Zhang2024ControlVideo}, long-duration generation~\cite{li2025longdiff,tan2024videoinf}, and physical dynamics simulation~\cite{montanaro2024motioncraft,yang2025vlipp}. Towards an immersive experience, researchers increasingly focus on integrating synchronized original audio~\cite{low2025ovi, hacohen2026ltx2, team2026mova}.

\begin{figure}[t]
\centerline{
\includegraphics[width=1.0\columnwidth]{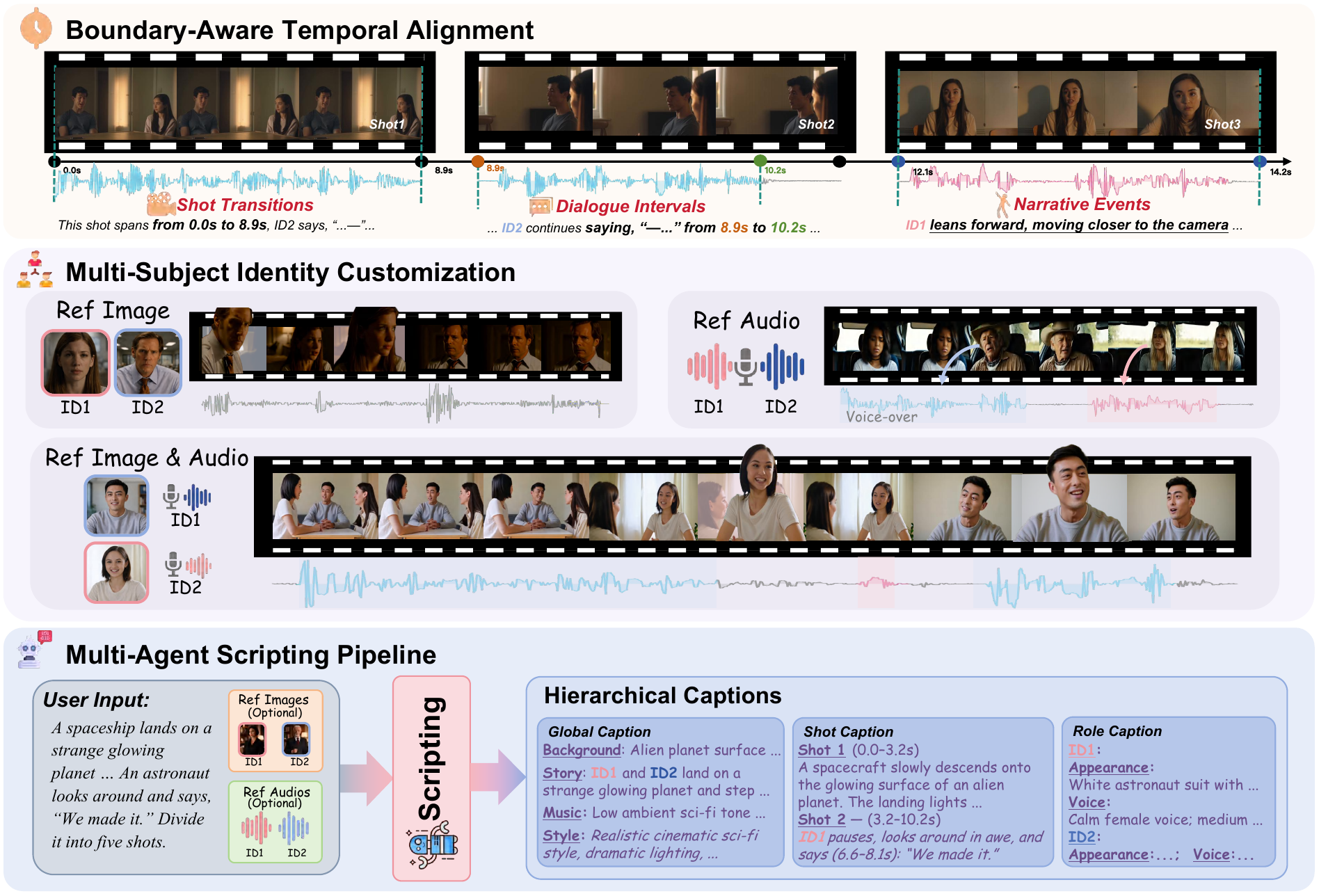}
}
\caption{
Illustration of our MAVIN framework.
\textbf{First row:} Leveraging boundary-aware attention, MAVIN enables precise temporal alignment for shot transitions, dialogue intervals, and narrative events.
\textbf{Second row:} Through ID-aware propagation, MAVIN allows users to customize multiple subjects via reference images and audio, maintaining identity consistency across complex cinematic narratives.
\textbf{Third row:} With the multi-agent scripting pipeline, MAVIN translates user inputs into hierarchical captions, providing decoupled narrative semantics for multi-shot audio-visual generation.
}

\label{fig:teaser}
\end{figure}

However, these clip-level models rely solely on a single text prompt for generation, rendering them impractical for real-world creative workflows (\eg, professional multi-shot filmmaking). In practice, creators depend on storyboards and structured scripts to precisely dictate the video content. To further empower filmmakers, recent efforts have focused on developing agents for story planning~\cite{wu2025automatedmovieagent} and synthesizing keyframes for shot transitions~\cite{he2025cut2next, zhang2025stage}, aiming to align with traditional human filmmaking workflows.

Despite the stronger controllability offered by these multi-shot video generation models, previous approaches primarily focus on the video-only modality, struggling with the challenges of audio-visual joint generation across complex narrative timelines for immersive filmmaking:
\textit{(i) Temporal misalignment}: Without explicit temporal boundaries, models struggle to synchronize specific visual elements (\eg, a character) with their audio events (\eg, speech) at the correct timestamps.
\textit{(ii) Limited controllability}: While current models effectively specify the visual appearance of subjects, their audio specifications (\eg, vocal timbre) remain largely unexplored, especially in multi-subject scenarios.
\textit{(iii) Incomplete scripting}: Prior scripting agents often omit audio-related narratives (\eg, human speech and sound effects), lacking time-stamped audio cues to guide the joint generation process.

To address these challenges, we propose \textbf{MAVIN}, the first framework for \textbf{M}ulti-Shot \textbf{A}udio-\textbf{V}isual generation with Custom\textbf{I}zed \textbf{N}arrative control. 
We build our framework upon OVI~\cite{low2025ovi} to leverage its priors for joint audio-visual representation and feature alignment. 
To avoid temporal misalignment, we propose boundary-aware attention, which partitions the latent space and routes corresponding tokens to strictly render audio and visual elements within their respective temporal boundaries (\cref{fig:teaser}, first row).
We further design an ID-aware propagation mechanism to accurately customize user-intended subjects, where ID tokens specify the target characters, and the audio-visual token context is leveraged to control the planned audio-visual attributes (\cref{fig:teaser}, second row). 
Finally, a lightweight multi-agent scripting pipeline is introduced to translate free-form prompts into comprehensive audio-visual narratives, providing temporal and structural cues for the joint generation process (\cref{fig:teaser}, third row).

To facilitate model training and evaluation, we introduce \textbf{MAVINSet}, a multi-shot audio-visual dataset with hierarchical captions, constructed from diverse datasets.
The training set comprises 800K samples, where each sample ranges from 3--15 seconds in duration and contains 1--6 shots, rendered at 24 fps and 480p resolution.
We further provide comprehensive hierarchical annotations at the global, shot, and role levels. These annotations explicitly specify the overall storyline, plot progression, dialogue intervals, action events, character appearance, and vocal timbre.
Finally, we establish an evaluation set of 1K manually verified samples to benchmark relevant methods.

Our contributions can be summarized as follows:
\begin{itemize}
    \item We propose the first framework for multi-shot audio-visual generation with customized narrative control, and introduce a multi-agent scripting pipeline to provide comprehensive audio-visual narratives as guidance.
    \item We design boundary-aware attention to render audio-visual elements within their respective temporal boundaries, and present an ID-aware propagation mechanism to specify target characters with planned audio-visual attributes.
    \item We construct a multi-shot audio-visual dataset with customized narrative annotations to facilitate model training, and establish a manually verified benchmark to rigorously evaluate the performance of relevant methods.
\end{itemize}

\section{Related Work}

\subsection{Joint Audio-Visual Generation}
Early audio-visual generation predominantly followed cascaded pipelines (\eg, video-to-audio~\cite{zhou2018visualtosound,luo2023difffoley} or audio-to-video~\cite{chatterjee2020sound2sight}), suffering from accumulated cross-modal errors and weak temporal synchronization.
Recent research has increasingly shifted toward joint audio-visual generation. Beyond early coupled-diffusion formulations such as MM-Diffusion~\cite{Ruan2023MMDiffusion}, methods like Seeing and Hearing~\cite{xing2024seeing} bridge pretrained audio and video generators through latent alignment. Concurrently, MM-LDM~\cite{Sun2024MMLDM} and AV-DiT~\cite{Wang2024AVDiT} move toward more unified latent and transformer-based architectures. More recent DiT-based systems, including UniForm~\cite{zhao2025uniform}, JavisDiT~\cite{liu2025javisdit}, JavisDiT++~\cite{liu2026javisditplusplus} and OVI~\cite{low2025ovi}, further strengthen cross-modal fusion and fine-grained synchronization through shared or twin-backbone designs. 
Adjacent cross-modal works such as AV-Link~\cite{haji2025avlink}, MMAudio~\cite{cheng2025mmaudio}, MTV~\cite{weng2025mtv}, and VinTAGe~\cite{Kushwaha2025VinTAGe} further demonstrate the importance of temporally aligned multimodal conditioning for synchronized audio-visual generation.
However, these methods largely assume a single contiguous event guided by a global prompt condition. 
Consequently, they still offer limited support for narrative-level planning and multi-shot semantic transitions.

\subsection{Multi-Shot and Narrative Video Generation}
To move beyond single-event clips, recent video generation research has begun to explicitly model multi-scene and long-horizon narratives. Planning-oriented methods such as VideoDirectorGPT~\cite{Lin2023VideoDirectorGPT}, VideoStudio~\cite{Long2024VideoStudio}, and STAGE~\cite{zhang2025stage} leverage LLMs, spatial layouts, storyboards, or reference images to decompose a global prompt into multiple scenes while preserving cross-scene consistency.
Hierarchical frameworks like MovieDreamer~\cite{zhao2024moviedreamer} further combine long-range autoregressive planning with diffusion rendering for coherent storytelling. At the generation level, StreamingT2V~\cite{Henschel2025StreamingT2V} propagates context across extended sequences, while multi-shot models such as ShotAdapter~\cite{Kara2025ShotAdapter} and EchoShot~\cite{wang2025echoshot} introduce explicit shot-aware conditioning to improve cross-shot controllability and identity preservation. Notably, Mind the Time~\cite{wu2025mindthetime} utilizes temporally localized captions with explicit start and end timestamps, enabling fine-grained control over the duration and ordering of multiple events. 
Nevertheless, these approaches remain fundamentally video-only, neglecting audio specifications (\eg, vocal timbre) for joint audio-visual generation.

\subsection{Controllable Video Generation}
Video controllability has rapidly evolved from basic structural conditioning to richer motion, camera, and identity control. Earlier ControlVideo~\cite{Zhang2024ControlVideo}, MotionCtrl~\cite{Wang2024MotionCtrl}, and Generative Rendering~\cite{cai2024generativerendering} focus on injecting structural, trajectory, or 4D guidance. Subsequent research extends this paradigm toward stronger motion and subject control. Frameworks including MotionBooth~\cite{Wu2024MotionBooth}, Motion Prompting~\cite{Geng2025MotionPrompting}, AnimateAnything~\cite{Lei2025AnimateAnything}, and MagicMotion~\cite{li2025magicmotion} support fine-grained object and camera motion specification under increasingly flexible control formats. In parallel, identity-oriented generation has advanced through multi-subject personalization methods such as Video Alchemist~\cite{chen2025videoalchemist}, improving visual subject fidelity in multi-entity scenes. Adjacent audio-aware editing and control methods further explore language-based colorization with audio alignment~\cite{chang2026lvocal}, audio-synchronized instance editing~\cite{zheng2026aviedit}, and instruction-guided joint audio-video editing~\cite{zheng2026instructav2av}.
Since existing models primarily customize the visual appearance, we propose MAVIN for audio-visual customized narrative, overcoming temporal misalignment, limited controllability, and incomplete scripting.

\begin{figure}[t]
\centerline{
\includegraphics[width=1.0\columnwidth]{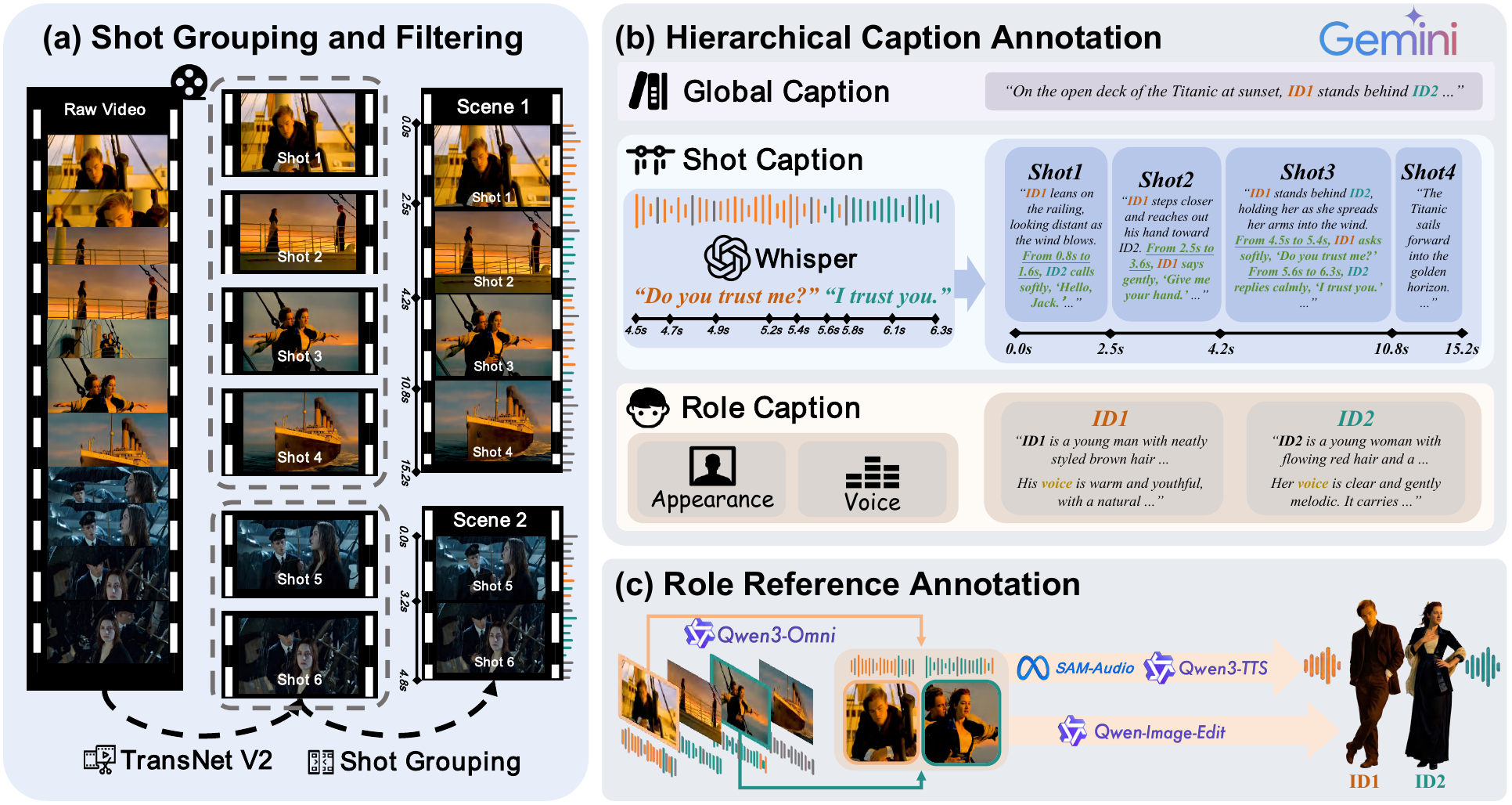}
}
\caption{Overview of our automatic data curation pipeline. 
\textbf{(a) Shot grouping and filtering.} We collect videos from diverse open-source datasets and online platforms. Raw videos are segmented into shots, grouped into multi-shot cinematic clips, and filtered for audio-visual aesthetics and clarity.
\textbf{(b) Hierarchical caption annotation.} We generate multi-level captions to capture global storylines, shot-level events, and role-level descriptions, alongside dialogue boundaries with decisecond-level timestamps.
\textbf{(c) Role reference annotation.} To establish consistent identity anchors, we extract audio-visual cues for each role and synthesize perturbed reference images and audio, decoupling identity from exact replication.}

\label{fig:dataset_pipeline}
\end{figure}

\section{Dataset}

Existing audio-visual datasets~\cite{Chen2020Vggsound, gemmeke2017audioset} mainly focus on single-shot events without structural annotations, leaving a data gap for complex storytelling. To train and evaluate MAVIN and establish a new benchmark for multi-shot generation, we propose MAVINSet, a large-scale dataset annotated with explicit temporal boundaries, dialogue intervals, and multiple audio-visual role references.

\noindent \textbf{Data sources.} To ensure both realism and diversity, \textsc{MAVINSet} is built upon diverse in-the-wild resources, including Condensed Movies~\cite{bain2020condensed}, Short-Films-20K~\cite{ghermi2025longstoryshortstorylevel}, MovieBench~\cite{Wu2025MovieBench}, VGGSound~\cite{Chen2020Vggsound}, and OpenHumanVid~\cite{Li2025OpenHumanVid} datasets, and publicly available online videos on YouTube.

\noindent\textbf{Shot unit grouping.}
As shown in Fig.~\ref{fig:dataset_pipeline} (a), we process raw videos by enforcing a minimum 480p resolution, removing black borders, and standardizing to 24 fps. We use TransNet V2~\cite{TransNetv2} for shot boundary detection and apply scene consistency representation learning~\cite{Wu2022Scene} to group shots belonging to the same scene segment. Each grouped unit includes 1--6 distinct shots spanning 3--15 seconds.

\noindent\textbf{Audio-visual quality filtering.}
We evaluate grouping candidates using the optical flow from RAFT~\cite{teed2020raft} to discard near-static segments. We then enforce perceptual quality thresholds across modalities, assessing video aesthetics with an aesthetic predictor~\cite{schuhmann2022improvedaesthetic} and audio quality with Audiobox-Aesthetics~\cite{tjandra2025audiobox}. To ensure clear identity representation, we use Qwen3-Omni~\cite{xu2025qwen3omni} and SAM2~\cite{ravi2024sam2} to discard clips with more than three individuals. We also apply Scribe~\cite{elevenlabs2026scribev2} for speaker diarization, excluding clips with interfering multi-speaker audio.

\noindent\textbf{Hierarchical caption annotation.}
As shown in Fig.~\ref{fig:dataset_pipeline} (b), we employ Gemini-2.5~\cite{comanici2025gemini25} to generate hierarchical captions, where the global level captures the overall storyline and plot progression, the role level details character appearance and vocal timbre, and the shot level describes action events and dialogue intervals. Notably, we additionally utilize Whisper~\cite{openaiwhisper} to extract timestamps for precise dialogue boundaries. If a dialogue extends beyond a single shot, we append a ``—'' to indicate continuation.

\noindent\textbf{Role reference annotation.}
As shown in Fig.~\ref{fig:dataset_pipeline} (c), for each multi-shot clip group, we extract visual and audio cues using Qwen3-Omni~\cite{xu2025qwen3omni} and SAM-Audio~\cite{shi2025samaudio}. We further synthesize perturbed visual references using Qwen-Image-Edit~\cite{wu2025qwenimagetechnicalreport} for altered poses or actions, and audio references using Qwen3-TTS~\cite{hu2026qwen3tts} for varying spoken content. This decouples identity from exact replication, enabling the model to generate dynamic actions and speech strictly guided by the narrative script.

\noindent\textbf{Manual benchmark verification.}
We create a 1K-sample high-fidelity benchmark from MAVINSet for evaluation. 
To prevent data leakage, we strictly enforce a raw-video-level disjoint split between the training set and evaluation benchmark.
We conduct manual verification to discard samples exhibiting identity drift, temporal misalignment, or structural inconsistencies that are beyond the reach of automatic metrics. We further recruit two independent annotators to assess whether each sample should be preserved. This process yielded a high inter-annotator agreement (Cohen's $\kappa = 0.77$), strongly demonstrating the benchmark's reliability.

\begin{figure}[t]
\centerline{
\includegraphics[width=1.0\columnwidth]{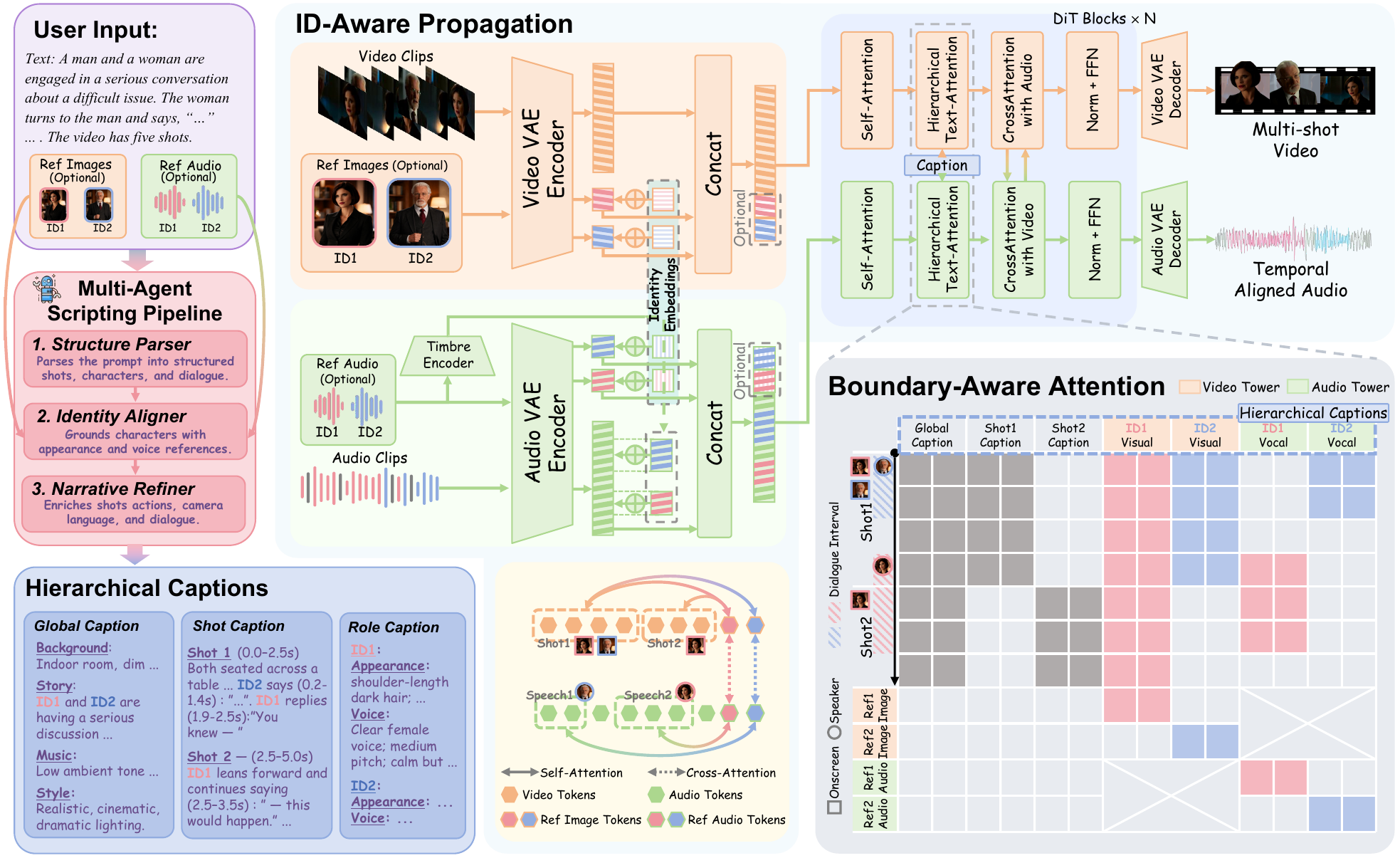}
}

\caption{
Given free-form user inputs \textbf{(purple block)}, we first employ a multi-agent scripting pipeline \textbf{(red block)} to transform them into hierarchical captions \textbf{(blue block)}, detailing complex narrative timelines with precise shot boundaries and dialogue intervals.  
Within the ID-aware propagation, input video clips and optional image references are compressed by a Video VAE into latent tokens \textbf{(orange block)}.
The audio tower follows a similar process, but its optional audio references are additionally processed by a pretrained timbre encoder to extract clean timbre embeddings \textbf{(green block)}.
After concatenating with learnable identity embeddings, these tokens are fed into a dual-tower diffusion Transformer \textbf{(cyan block)}. 
We adapt the self-attention and text-attention layers to be compatible with customized identities \textbf{(yellow block)}.
Furthermore, we integrate boundary-aware attention into these text-attention layers to enforce strict temporal alignment across complex multi-shot narratives \textbf{(gray block)}. 
Finally, the VAE decoders produce the synchronized multi-shot video and audio results.
}

\label{fig:pipeline}
\end{figure}

\section{Method}

In this section, we detail our proposed \textbf{MAVIN} framework.
We first provide an overview of the model architecture and optimization process (\cref{sec:overview}). 
Next,  we introduce the boundary-aware attention mechanism to address temporal misalignment across complex narratives (\cref{sec:boundary_attention}).
To achieve precise identity and vocal timbre customization, we further detail the ID-aware propagation strategy (\cref{sec:id_propagation}). 
Finally, we describe the multi-agent scripting pipeline to transform free-form user prompts into the hierarchical captions (\cref{sec:scripting_pipeline}). The overall pipeline is shown in \cref{fig:pipeline}.

\subsection{Overview}
\label{sec:overview}

In this section, we present an overview of our MAVIN framework.

\noindent\textbf{Audio and video compression.}
MAVIN employs pretrained Variational Autoencoders (VAEs) to compress audio and video modalities into their latent spaces. 
Given a video $\mathbf{x}^v$ and its accompanying audio $\mathbf{x}^a$, the respective VAE encoders compress them into compact latent codes $\mathbf{z}_0^v = E_v(\mathbf{x}^v)$ and $\mathbf{z}_0^a = E_a(\mathbf{x}^a)$.

\noindent\textbf{Model architecture.}
Following previous works~\cite{low2025ovi,hacohen2026ltx2, team2026mova}, we design our prediction network as a dual-tower diffusion Transformer framework.
Given compressed latent codes $\mathbf{z}_t^v$ and $\mathbf{z}_t^a$ at timestep $t$, hierarchical captions $\mathcal{C}^\mathrm{T}$, and optional role references $\mathcal{C}^\mathrm{R}$, the framework is equipped with stacked Transformer blocks. These include self-attention layers to extract audio-visual features, hierarchical text-attention layers to decouple multi-level narrative semantics, and cross-attention layers for fine-grained audio-visual synchronization. The framework conducts joint prediction of the audio-visual velocity fields.

\noindent\textbf{Optimization process.}
The framework is optimized via flow matching~\cite{lipman2022flowmatching}, where independent Gaussian priors $\boldsymbol{\epsilon}^v, \boldsymbol{\epsilon}^a \sim \mathcal{N}(0, I)$ are transformed into the target audio-visual latent codes. The linear interpolation paths are defined as $\mathbf{z}_t^v = (1 - t)\mathbf{z}_0^v + t\boldsymbol{\epsilon}^v$ and $\mathbf{z}_t^a = (1 - t)\mathbf{z}_0^a + t\boldsymbol{\epsilon}^a$ for $t \sim \mathcal{U}(0,1)$. Specifically, the model is optimized to regress the vector fields using the following objective:
\begin{align}
    \mathcal{L}_\mathrm{FM} = \mathbb{E}_{t,\mathbf{z}_0^v,\mathbf{z}_0^a,\boldsymbol{\epsilon}^v,\boldsymbol{\epsilon}^a} 
    \Big[ & \left\|v_\theta^v(\mathbf{z}_t^v, t, \mathcal{C}^\mathrm{T}, \mathcal{C}^\mathrm{R}) - (\boldsymbol{\epsilon}^v - \mathbf{z}_0^v)\right\|_2^2 + \nonumber \\
    & \left\|v_\theta^a(\mathbf{z}_t^a, t, \mathcal{C}^\mathrm{T}, \mathcal{C}^\mathrm{R}) - (\boldsymbol{\epsilon}^a - \mathbf{z}_0^a)\right\|_2^2 \Big],
\end{align}
where $v_\theta^v(\dots)$ and $v_\theta^a(\dots)$ are the velocity fields predicted by the video and audio towers, respectively.
During inference, the network synthesizes the final synchronized multi-shot audio-visual sequence by solving the probability flow Ordinary Differential Equation (ODE) defined by the predicted velocity fields.

\subsection{Boundary-Aware Attention}
\label{sec:boundary_attention}

Clip-level video generation models typically adopt a single text prompt to guide the generation process, where all text tokens are visible to all latent positions. However, for multi-shot generation, this implicit cross-attention inevitably leads to semantic leakage from one shot into another. 
To address this, we adopt hierarchical captions to decouple global, shot, and role-level narrative semantics, while routing tokens to render elements within their respective temporal boundaries, as presented in~\cref{fig:pipeline} bottom right.

\noindent\textbf{Hierarchical caption formulation.} 
For a multi-shot narrative, we formulate the input caption into a hierarchical format with three levels:
\textit{(i) Global caption} ($\mathcal{C}^\textrm{global} = \{\mathbf{g}\}$) presents the overall storyline and plot progress across the entire generated sequence $[0, T]$, where $\mathbf{g}$ denotes the global narrative context;
\textit{(ii) Shot caption} ($\mathcal{C}^\textrm{shot}_{i} = \{(\mathbf{s}_i, \mathcal{T}^\mathrm{s}_i)\}$) guides dialogue intervals and action events, strictly anchored to the defined timestamps of the $i$-th cinematic shot, where $\mathbf{s}_i$ is the description of shot $i$ within the shot interval $\mathcal{T}^\mathrm{s}_i$;
\textit{(iii) Role caption} ($\mathcal{C}^\textrm{role}_{j} = \{(\mathbf{d}_j^\textrm{v}, \mathbf{d}_j^\textrm{a}, \mathcal{T}^\mathrm{r}_j)\}$) provides fine-grained identity anchoring. The visual appearance descriptor $\mathbf{d}_j^\textrm{v}$ is activated only when role $j$ is present in the shot, while the audio specification descriptor $\mathbf{d}_j^\textrm{a}$ is presented within its dialogue intervals $\mathcal{T}^\mathrm{r}_j$. As a result, the hierarchical caption $\mathcal{C}$ is formulated as:
\begin{equation}
\label{eq:task_formulation}
\mathcal{C} = \Big\{ \{\mathbf{g}\},\; \{(\mathbf{s}_i, \mathcal{T}^\mathrm{s}_i)\}_{i=1}^{S},\; \{(\mathbf{d}_j^\mathrm{v}, \mathbf{d}_j^\mathrm{a}, \mathcal{T}^\mathrm{r}_j)\}_{j \in \mathcal{R}} \Big\},
\end{equation}
where $S$ denotes the number of shots and $\mathcal{R}$ specifies the set of characters.

\noindent\textbf{Latent boundary partition.}
Since videos and audio are presented with distinct physical natures, we propose an asymmetric boundary partition strategy for the latent tokens to construct overlapping segments. 
Since video frames are continuous, we partition its latent codes $\mathbf{Z}_i^\mathrm{v}$ from the video latent space $\mathbf{Z}^\mathrm{v}$ according to shot transitions as $\mathbf{Z}_i^\mathrm{v} = \left\{ \mathbf{z}_k^\mathrm{v} \mid \tau(k) \in \mathcal{T}^\mathrm{s}_i \right\}$.
In contrast, since character speech is typically sparse and event-driven, we partition its latent codes $\mathbf{Z}_{j}^\mathrm{a}$ from the audio latent space $\mathbf{Z}^\mathrm{a}$ based on the dialogue interval $\mathcal{T}^\mathrm{r}_j$ of role $j$ as $\mathbf{Z}_j^\mathrm{a} = \left\{ \mathbf{z}_k^a \mid \tau(k) \in \mathcal{T}^\mathrm{r}_j \right\}$.

\noindent \textbf{Boundary-aware token routing.} 
To strictly generate multi-shot videos following this structured input, we replace the standard cross-attention layer with the hierarchical text-attention layer. 
Denoting a binary routing matrix $\mathbf{M} \in \{0, 1\}^{N \times L}$, where $N$ is the latent sequence length and $L$ is the text token length, the attention calculation is reformulated as:
\begin{equation}
    \mathbf{Z} = \mathrm{softmax}\left(\frac{\mathbf{Q} \mathbf{K}_{\mathcal{C}}^\top}{\sqrt{d}} + \log \mathbf{M}\right) \mathbf{V}_{\mathcal{C}},
\end{equation}
where $\mathbf{Q}$ is derived from the latent codes, $\mathbf{K}_{\mathcal{C}}$ and $\mathbf{V}_{\mathcal{C}}$ are projected from the hierarchical captions $\mathcal{C}$, $d$ is the head dimension, and the routing matrix $\mathbf{M}$ computes visibility dynamically based on the partitioned temporal boundaries. For a video latent token $k$ at time $\tau(k)$ and a textual token $c$ from the hierarchical caption $\mathcal{C}$, the visibility is defined as:
\begin{equation}
     \mathbf{M}^\mathrm{v}(k, c) = 
    \begin{cases} 
        1 & \text{if } c \in \mathcal{C}^\textrm{global}, \\ 
        1 & \text{if } c \in \mathcal{C}^\textrm{shot}_{i} \land \tau(k) \in \mathcal{T}^\mathrm{s}_{i}, \\ 
        1 & \text{if } c \in \mathcal{C}^\textrm{role}_{j} \land \tau(k) \in \mathcal{T}^\mathrm{s}_{i} \land j \in \mathcal{R}_i, \\ 
        0 & \text{otherwise},
    \end{cases}
\end{equation}
where $\mathcal{R}_i$ means the set of characters presented in $i$-th shot.
Similarly, for audio latent codes, the role-level audio specification descriptor $\mathbf{d}_j^\mathrm{a}$ is routed to $\mathbf{M}^\mathrm{a}(k, c) = 1$ only if $\tau(k) \in \mathcal{T}^\mathrm{r}_j$. 
Finally, we denote the boundary-aware attention mechanism as the collection of latent boundary partition and boundary-aware token routing, enabling rendering audio and visual elements within their respective temporal boundaries.

\subsection{ID-Aware Propagation}
\label{sec:id_propagation}

To improve the controllability of audio-visual generation and enable customized generation for multi-subject scenarios (\ie, binding roles to specific visual appearances and vocal timbres), we introduce identity embeddings for appearance and timbre modulation, and identity-aware injection for customized token conditioning during generation, as presented in~\cref{fig:pipeline} middle.

\noindent\textbf{Identity embeddings.}
We design identity embeddings for the cases where the user specifies customized roles, each accompanied by optional reference images and audio clips. 
In this case, for each customized role $j$, we feed its reference inputs into its respective VAE encoders to obtain tokens representing the visual appearance $\mathbf{Z}^{\mathrm{v},\mathrm{ref}}_j$ and audio specifications $\mathbf{Z}^{\mathrm{a},\mathrm{ref}}_j$. 
Since reference audio inherently entangles timbre with semantics, emotion, and prosody, we additionally employ a pretrained timbre encoder~\cite{ju2024naturalspeech} to extract a clean timbre embedding $\mathbf{e}_j$. 
Before fusion, we project $\mathbf{e}_j$ to the audio-token hidden dimension and broadcast it to match the shape of $\mathbf{Z}^{\mathrm{a},\mathrm{ref}}_j$.
To explicitly prevent identity collapse among similar roles, we modulate these extracted features with learnable role ID embeddings $\mathbf{p}_j^\mathrm{v}, \mathbf{p}_j^\mathrm{a} \in \mathbb{R}^d$ as:
\begin{equation}
\hat{\mathbf{Z}}^{\mathrm{v},\mathrm{ref}}_j = \mathbf{Z}^{\mathrm{v},\mathrm{ref}}_j + \mathbf{p}_j^{\mathrm{v}}, \qquad \hat{\mathbf{Z}}^{\mathrm{a},\mathrm{ref}}_j = \mathbf{Z}^{\mathrm{a},\mathrm{ref}}_j + \mathbf{p}_j^{\mathrm{a}} + \mathbf{e}_j.
\end{equation}
These augmented identity anchors serve as customized tokens and are concatenated with their respective target latent codes ($\mathbf{Z}_j^\mathrm{a}$ and $\mathbf{Z}_i^\mathrm{v}$), anchoring identity priors to the target characters before the diffusion Transformer. If no customized role is specified, this concatenation is omitted.

\noindent\textbf{Identity-aware mask.}
To prevent the reference images and audio from interfering with the original audio-visual generation process, we apply an additional visibility mask within each attention layer during customization.
\textit{(i) Video self-attention layer:} Video tokens $\mathbf{Z}_i^\mathrm{v}$ for the $i$-th shot can only interact with themselves, or with the visual anchors $\hat{\mathbf{Z}}^{\mathrm{v},\mathrm{ref}}_j$ of the roles $j$ present in the same shot.
\textit{(ii) Audio self-attention layer:} Audio tokens $\mathbf{Z}_{j}^\mathrm{a}$ can only attend to themselves, or the vocal anchors $\hat{\mathbf{Z}}^{\mathrm{a},\mathrm{ref}}_j$ speaking within their timestamp intervals $\mathcal{T}^\mathrm{r}_j$.
\textit{(iii) Text-attention layer:} Role-level text prompts strictly attend to their corresponding visual and vocal anchors (\ie, $\mathcal{C}_j^\mathrm{role} \leftrightarrow \{\hat{\mathbf{Z}}^{\mathrm{v},\mathrm{ref}}_j, \hat{\mathbf{Z}}^{\mathrm{a},\mathrm{ref}}_j\}$) to propagate accurate role identity.
\textit{(iv) Cross-attention layer:} Only tokens belonging to the visual and vocal anchors of the same role are permitted to interact ($\hat{\mathbf{Z}}^{\mathrm{a},\mathrm{ref}}_j \leftrightarrow \hat{\mathbf{Z}}^{\mathrm{v},\mathrm{ref}}_j$).
With this carefully designed masking strategy, our model is elegantly adapted for audio-visual role customization.

\subsection{Multi-Agent Scripting Pipeline}
\label{sec:scripting_pipeline}

In our multi-shot audio-visual generation model, the narrative script with hierarchical captions directly decides the final generation quality. Since user-provided prompts $\mathcal{P}$ are typically free-form, lacking fine-grained details for scenario descriptions or role traits, we introduce a lightweight multi-agent scripting pipeline with an off-the-shelf LLM~\cite{xu2025qwen3omni} to transform $\mathcal{P}$ into the hierarchical captions $\mathcal{C}$, as presented in~\cref{fig:pipeline} left.

Specifically, we implement this caption augmentation through three specialized agents. 
First, the \textit{structure parser} takes the user-provided prompt $\mathcal{P}$ and utilizes predefined templates to extract global scene attributes, decomposing each shot boundary $\{\mathcal{T}^\mathrm{s}_i\}_{i=1}^S$, identifying the discrete set of characters $\mathcal{R}$, and extracting their dialogue intervals $\{\mathcal{T}^\mathrm{r}_j\}_{j \in \mathcal{R}}$. 
Next, the \textit{identity aligner} grounds these roles by integrating optional user-provided visual and audio references. Finally, the \textit{narrative refiner} enriches the shot-level action events, camera language, and dialogues, producing the final captions $\mathcal{C} = \big\{ \mathcal{C}^\mathrm{global}, \{\mathcal{C}^\mathrm{shot}_i\}_{i=1}^S, \{\mathcal{C}^\mathrm{role}_j\}_{j\in \mathcal{R}} \big\}$, which are tailored for the formulation of our generative framework.

\section{Experiments}

\subsection{Implementation Details}
\noindent\textbf{Basic setup.}
We adopt OVI~\cite{low2025ovi} as the backbone for our joint audio-visual generation, supporting variable clip lengths of 3 to 15 seconds at 24 fps with 480p resolution. Training is performed within a joint latent space encoded by 3D video and audio VAEs.
To handle variable sequence lengths, we employ dynamic resolution bucket-cropping and precise temporal alignment strategies.

\noindent\textbf{Multi-stage progressive training.}
To prevent optimization conflicts caused by diverse conditioning signals (\ie, multi-level narrative semantics, customized visual appearances and vocal timbres), we adopt a progressive alignment training schedule. We sequentially optimize for text-driven generation (Stage 1), introduce single-modality identity anchors for visual and audio references (Stage 2), and finally perform joint training integrating all conditions (Stage 3).

\begin{table}[t]
  \centering
  \caption{Quantitative comparison of our framework with state-of-the-art audio-visual generation methods. Throughout the paper, $\uparrow$ ($\downarrow$) indicates higher (lower) is better. The best results are highlighted with \textbf{bold}, and `-' indicates the metric is inapplicable.}
  \label{tab:main_results}
  \resizebox{\textwidth}{!}{
  \begin{tabular}{l | cc | ccc | ccc | ccccc}
    \toprule
    \multirow{2}{*}{\textbf{Method}} 
    & \multicolumn{2}{c|}{\textbf{Quality}} 
    & \multicolumn{3}{c|}{\textbf{Semantics}} 
    & \multicolumn{3}{c|}{\textbf{AV-Alignment}} 
    & \multicolumn{5}{c}{\textbf{Multi-shot Consistency}} \\
    \cmidrule(lr){2-3} \cmidrule(lr){4-6} \cmidrule(lr){7-9} \cmidrule(lr){10-14}
    & FVD $\downarrow$ & FAD $\downarrow$ 
    & TVS $\uparrow$ & TAS $\uparrow$ & WER $\downarrow$
    & Sync $\uparrow$ & AV-IB $\uparrow$ & TAMS $\uparrow$ 
    & SC $\uparrow$ & BC $\uparrow$ & CISC $\uparrow$ & BISC $\uparrow$ & STA $\uparrow$ \\
    \midrule
    \multicolumn{14}{l}{\textit{- Cascaded Generation Models: Multi-shot T2V + V2A}} \\
    VideoGen-of-Thought~\cite{zheng2024videogenofthought} & 418.7 & 13.2 & 0.1537 & 0.1939 & - & 2.278 & 0.184 & - & 0.9593 & 0.9578 & {0.6195} & 0.7802 & - \\
    MovieAgent~\cite{wu2025automatedmovieagent}          & 395.3 & 15.9 & 0.1271 & 0.1995 & - & 1.913 & 0.179 & - & {0.9656} & 0.9298 & 0.4872 & 0.6744 & - \\
    EchoShot~\cite{wang2025echoshot}            & 276.3 & 12.3 & 0.2072 & 0.2191 & - & 2.095 & 0.207 & - & 0.9532 & 0.9572 & 0.6104 & 0.7599 & {0.6002} \\
    CineTrans~\cite{wu2025cinetrans}           & {254.2} & 11.9 & 0.2013 & 0.2098 & - & 1.892 & 0.201 & - & 0.9425 & 0.9543 & 0.6073 & 0.7603 & 0.5820 \\
    IC-LoRA~\cite{huang2024iclora} + Wan~\cite{wan2025wan}      & 268.5 & 12.7 & {0.2193} & 0.2012 & - & 2.103 & 0.192 & - & 0.9512 & 0.9626 & 0.5286 & 0.7519 & - \\
    StoryDiff.~\cite{zhou2024storydiffusion} + Wan~\cite{wan2025wan}   & 304.6 & 13.3 & 0.2124 & 0.1978 & - & 2.013 & 0.191 & - & 0.9487 & {0.9654} & 0.5613 & {0.7993} & - \\
    \midrule
    \multicolumn{14}{l}{\textit{- Joint Generation Models: T2AV}} \\
    JavisDiT~\cite{liu2025javisdit}            & 512.6 & 21.3 & 0.163 & 0.1937 & 0.305 & 3.734 & 0.239 & 0.1012 & 0.9512 & 0.9567 & - & - & - \\
    UniVerse-1~\cite{wang2025universe1}          & 356.9 & 11.2 & 0.182 & 0.1720 & 0.199 & 3.851 & 0.193 & 0.2892 & 0.9499 & 0.9601 & - & - & - \\
    OVI~\cite{low2025ovi}                 & 318.4 & {8.3} & 0.1983 & 0.2203 & {0.093} & 4.231 & {0.227} & 0.4432 & 0.9565 & 0.9578 & 0.6121 & 0.7789 & 0.4992 \\
    LTX-2~\cite{hacohen2026ltx2}               & 289.7 & 8.6 & 0.2001 & {0.2287} & 0.097 & {4.365} & 0.203 & {0.5343} & 0.9593 & 0.9553 & 0.6015 & 0.7812 & 0.5733 \\
    \midrule
    \textbf{Ours (MAVIN)}       & \textbf{231.6} & \textbf{6.8} & \textbf{0.2471} & \textbf{0.2392} & \textbf{0.048} & \textbf{6.032} & \textbf{0.263} & \textbf{0.8104} & \textbf{0.9695} & \textbf{0.9657} & \textbf{0.6319} & \textbf{0.8013} & \textbf{0.9897} \\
    \bottomrule
  \end{tabular}
  }
\end{table}

\subsection{Evaluation Metrics}
We comprehensively evaluate MAVIN from the following perspectives:

\noindent \textbf{Perceptual quality.} We measure the fidelity of generated video and audio using Fréchet Video Distance (FVD)~\cite{unterthiner2018fvd} and Fréchet Audio Distance (FAD)~\cite{kilgour2018FAD}.

\noindent \textbf{Semantic consistency.} We employ ViCLIP~\cite{wang2023internvid} and CLAP~\cite{elizalde2023clap} to measure Text-Video Similarity (TVS) and Text-Audio Similarity (TAS), respectively. Additionally, we compute the Word Error Rate (WER) evaluated via Whisper-large-v3~\cite{openaiwhisper} to assess the accuracy of the spoken words.

\noindent\textbf{Audio-visual alignment.} We assess lip synchronization via SyncNet~\cite{chung2016syncnet} and overall cross-modal semantic similarity using ImageBind~\cite{girdhar2023imagebind}.
We further introduce the Time Alignment Metric for Speech (TAMS) to quantify the synchronization between the synthesized speech and the designated temporal boundaries (\ie, dialogue intervals).

\noindent\textbf{Multi-shot consistency.}  
We compute Subject Consistency (SC) and Background Consistency (BC) following VBench~\cite{huang2024vbench} to assess intra-shot temporal stability.
Furthermore, we calculate Character and Background Inter-shot Consistency (CISC, BISC) scores utilizing ViCLIP~\cite{wang2023internvid} features.
Finally, we introduce Shot Transition Accuracy (STA) to evaluate the model's precision in executing shot transitions based on the narrative scripts.

\subsection{Comparison with State-of-the-Art Methods}
\noindent\textbf{Baselines.} 
We evaluate MAVIN against two primary paradigms of audio-visual generation on our 1K-sample high-fidelity benchmark. For a fair comparison, all models are evaluated without providing role references.

\noindent \textbf{Cascaded generation models.} 
We first evaluate multi-shot Text-to-Video (T2V) models (\ie, VideoGen-of-Thought~\cite{zheng2024videogenofthought}, MovieAgent~\cite{wu2025automatedmovieagent}, EchoShot~\cite{wang2025echoshot}, CineTrans~\cite{wu2025cinetrans}, IC-LoRA~\cite{huang2024iclora}, and StoryDiffusion~\cite{zhou2024storydiffusion}), followed by a Video-to-Audio (V2A) model (Hunyuan-Foley~\cite{shan2025hunyuanvideofoley}) to generate the accompanying audio tracks.
Notably, since IC-LoRA~\cite{huang2024iclora} and StoryDiffusion~\cite{zhou2024storydiffusion} only generate keyframes, we equip them with Wan2.2~\cite{wan2025wan} for generating the video frames.

\noindent \textbf{Joint generation models.}
We compare MAVIN with end-to-end joint Text-to-Audio-Visual (T2AV) generation models (\ie, JavisDiT~\cite{liu2025javisdit}, UniVerse-1~\cite{wang2025universe1}, OVI~\cite{low2025ovi}, and LTX-2~\cite{hacohen2026ltx2}), under the same inference settings as ours.

\noindent \textbf{Quantitative results.}
As shown in Tab.~\ref{tab:main_results}, MAVIN outperforms all state-of-the-art baselines across 13 metrics.
\textit{(i) Cascaded generation models.}
While these models~\cite{zheng2024videogenofthought,wu2025automatedmovieagent,wang2025echoshot,wu2025cinetrans,huang2024iclora,zhou2024storydiffusion} achieve strong visual consistency (SC, BC) via dedicated T2V networks, their decoupled pipeline restricts the downstream V2A model to rigidly fit the generated videos.
Since existing T2V models lack fine-grained lip motion, the downstream V2A models fail to generate intelligible speech, making WER and TAMS inapplicable.
\textit{(ii) Joint generation models.}
Existing joint models~\cite{liu2025javisdit,wang2025universe1,low2025ovi,hacohen2026ltx2} improve basic audio-visual alignment but lack explicit mechanisms for hard cuts and long-term spatial correlations. Consequently, they suffer from character and background drifting (lower CISC/BISC) and fail to execute precise scene transitions (STA $< 0.60$). In contrast, our framework leverages boundary-aware attention to achieve a remarkable STA of 0.9897 and a CISC of 0.6319. 
Additionally, our framework achieves a state-of-the-art TAMS score of 0.8104, effectively ensuring that characters speak strictly within their designated temporal boundaries.

\begin{figure}[t]
\centerline{
\includegraphics[width=1.0\columnwidth]{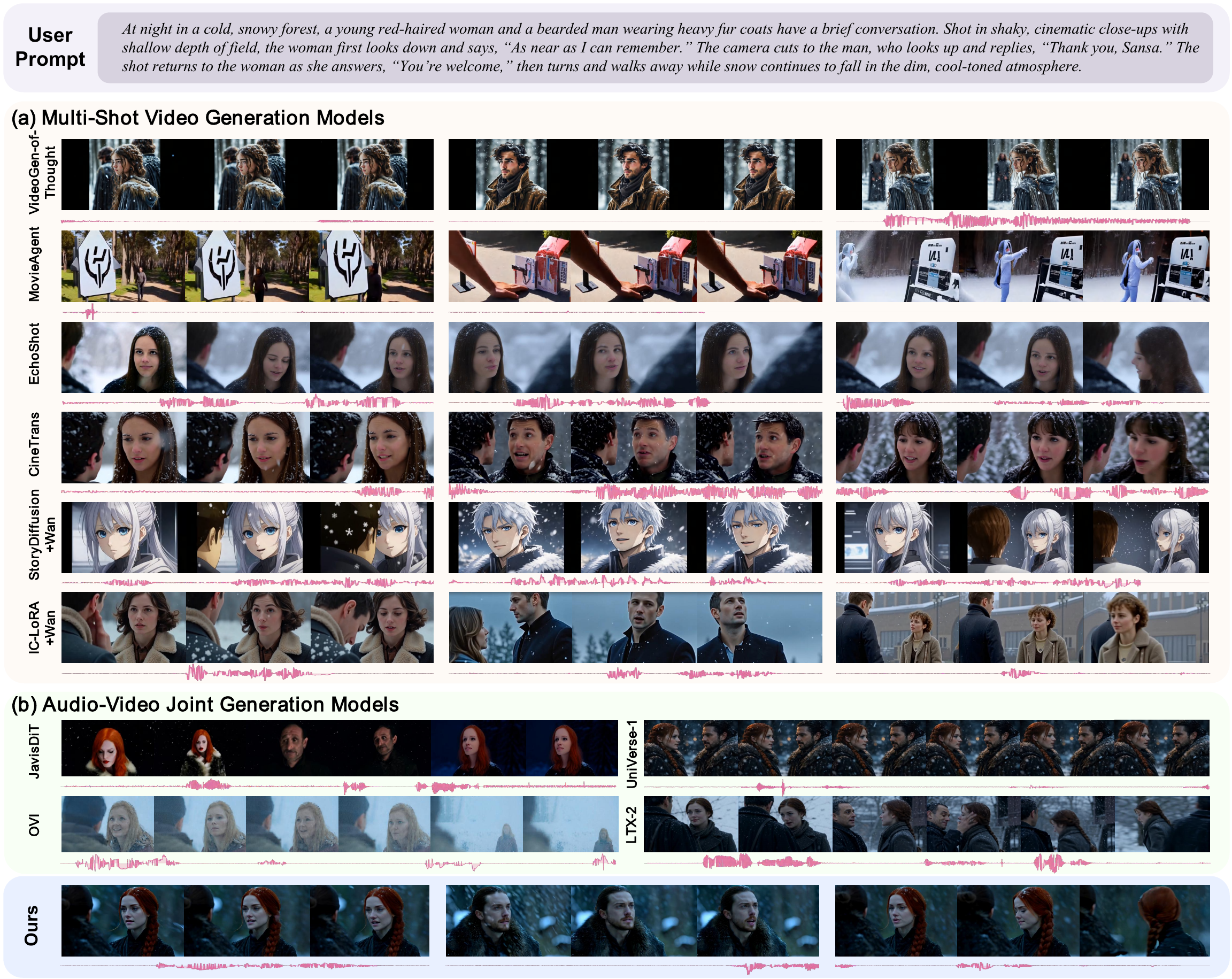}
}
\caption{Qualitative comparison with state-of-the-art relevant generation models. }

\label{fig:qualitative_comparison}
\end{figure}

\noindent \textbf{Qualitative results.}
In Fig.~\ref{fig:qualitative_comparison}, we present qualitative comparisons.
\textit{(i) Cascaded generation models} generate visually plausible individual shots but suffer from severe cross-modal misalignment, including lip synchronization failures and abrupt audio discontinuities at transition boundaries.
\textit{(ii) Joint generation models} improve audio-visual coherence, but most of them still struggle with spatiotemporal consistency for multi-shot scenes, leading to noticeable character drifting and background inconsistency after cinematic cuts.
Instead, our framework consistently preserves visual appearance and vocal timbre across shots with precise temporal alignment.

\subsection{Ablation Study}
\label{sec:ablation}
\begin{table}[t]
\centering
\caption{Quantitative ablation study. Removing modules consistently degrades quality. }
\label{tab:ablation_allmetrics}
\small
\resizebox{\textwidth}{!}{
\begin{tabular}{l | cc | ccc | ccc | ccccc}
\toprule
\multirow{2}{*}{Variant}
& \multicolumn{2}{c|}{Quality}
& \multicolumn{3}{c|}{Semantics}
& \multicolumn{3}{c|}{AV-Alignment}
& \multicolumn{5}{c}{Multi-shot Consistency} \\
\cmidrule(lr){2-3}\cmidrule(lr){4-6}\cmidrule(lr){7-9}\cmidrule(lr){10-14}

& FVD $\downarrow$ & FAD $\downarrow$
& TVS $\uparrow$ & TAS $\uparrow$ & WER $\downarrow$
& Sync $\uparrow$ & AV-IB $\uparrow$ & TAMS $\uparrow$
& SC $\uparrow$ & BC $\uparrow$ & CISC $\uparrow$ & BISC $\uparrow$ & STA $\uparrow$\\

\midrule

w/o BTR
& 268.4 & 7.6
& 0.2310 & 0.2214 & 0.062
& 5.830 & 0.241 & 0.582
& 0.9631 & 0.9592 & 0.5520 & 0.7410 & 0.5580 \\

w/o IM
& 256.9 & 7.3
& 0.2398 & 0.2301 & 0.055
& 5.970 & 0.252 & 0.787
& 0.9642 & 0.9617 & 0.5710 & 0.7680 & 0.8010 \\

w/o PTE
& 242.5 & 7.0
& 0.2453 & 0.2368 & 0.051
& 6.010 & 0.258 & 0.792
& 0.9681 & 0.9643 & 0.6240 & 0.7920 & 0.9460\\

w/o MAP
& 261.1 & 7.5
& 0.2336 & 0.2248 & 0.058
& 5.910 & 0.244 & 0.569
& 0.9653 & 0.9618 & 0.5900 & 0.7700 & 0.6230\\

\midrule
\textbf{MAVIN}
& \textbf{241.9} & \textbf{6.9}
& \textbf{0.2463} & \textbf{0.2391} & \textbf{0.049}
& \textbf{6.057} & \textbf{0.266} & \textbf{0.793}
& \textbf{0.9682} & \textbf{0.9653} & \textbf{0.6397} & \textbf{0.7996} & \textbf{0.9886}\\

\bottomrule
\end{tabular}
}
\end{table}

As presented in Tab.~\ref{tab:ablation_allmetrics}, we ablate key components of our framework to evaluate their individual contributions and report these quantitative results. Since some ablations target customized generation capabilities, we conduct these experiments with explicit video and audio references.

\noindent \textbf{W/o BTR (Boundary-aware Token Routing).} Removing BTR allows tokens to interact freely across designated temporal boundaries, leading to semantic leakage between shots. This uncontrolled interaction significantly degrades multi-shot consistency metrics (CISC, BISC, and STA).

\noindent \textbf{W/o IM (Identity-aware Mask).} Disabling IM allows identity anchors to interact freely across different shots and roles. This results in inter-subject crosstalk and identity blending in both visual appearance and vocal timbre, damaging character inter-shot consistency (CISC).

\noindent\textbf{W/o PTE (Pretrained Timbre Encoder).} Without the specialized PTE, the model struggles to decouple vocal timbre from other acoustic cues (\eg, linguistic semantics, emotion, and prosody). This entanglement impacts the overall cross-modal synchronization (Sync, AV-IB).

\noindent \textbf{W/o MAP (Multi-agent Scripting Pipeline).} Directly feeding free-form prompts without MAP fails to provide the decoupled narrative semantics. Consequently, the model struggles with executing precise scene transitions (STA) and maintaining temporal alignment (TAMS).

\subsection{User Study}
\noindent\textbf{Generation quality.}
To further assess subjective human preference, we conduct a user study across three key dimensions: 
\textit{(i)} \textbf{Audio-Visual Quality (AVQ)}, evaluating the perceptual fidelity of generated videos and their accompanying audio; 
\textit{(ii)} \textbf{Multi-shot Identity Consistency (MIC)}, measuring the stability of role attributes across shot boundaries without identity blending; and 
\textit{(iii)} \textbf{Audio/Video Narrative Adherence (ANA/VNA)}, assessing the precise alignment between generated audio/video events and the structured scripting.
In each experiment, participants are presented with generated audio-visual results from MAVIN and baseline methods, alongside the input free-form prompts. They are asked to select the result that best aligns with each evaluation dimension. We conduct the user study on Amazon Mechanical Turk (AMT) using 20 samples randomly selected from the MAVINSet dataset, with results polled from 25 volunteers. 
As shown in Tab.~\ref{tab:user_study}, our framework consistently achieves the highest preference scores across all dimensions, demonstrating superior human subjective performance.

\noindent \textbf{Agent quality.}
To evaluate whether the multi-agent scripting pipeline effectively produces reliable hierarchical captions for downstream inference, we conduct an additional user study to measure the alignment between generated captions and intended narrative semantics. 
We ask 25 volunteers to evaluate 20 random samples and categorize the generated caption quality as ``Failed'', ``Borderline'', ``Acceptable'', or ``Perfect''. The resulting distribution is 1.2\%, 4.2\%, 10.4\%, and 84.2\%, respectively. With 94.6\% of the ratings being ``Acceptable'' or higher, our scripting pipeline demonstrates strong robustness and reliability.

\begin{table}[tb]
  \centering
  \caption{Percentage (\%) of user preference in the subjective evaluation.}
  \label{tab:user_study}
  \small
  \setlength{\tabcolsep}{4pt}
  \resizebox{\linewidth}{!}{%
  \begin{tabular}{l cccc @{\hspace{10pt}} l cccc}
    \toprule
    \multicolumn{5}{c}{\textit{- Cascaded Generation Models: Multi-shot T2V + V2A}} &
    \multicolumn{5}{c}{\textit{- Joint Generation Models: T2AV}} \\
    \cmidrule(lr){1-5}\cmidrule(lr){6-10}
    Method & AVQ & MIC & ANA & VNA &
    Method & AVQ & MIC & ANA & VNA \\
    \midrule
    VideoGen-of-Thought~\cite{zheng2024videogenofthought} & 5.6 & 8.0 & 3.2 & 3.4 &
    JavisDiT~\cite{liu2025javisdit} & 3.4 & 3.0 & 2.0 & 2.2 \\
    MovieAgent~\cite{wu2025automatedmovieagent} & 4.8 & 4.6 & 2.6 & 2.8 &
    UniVerse-1~\cite{wang2025universe1} & 4.6 & 4.0 & 2.8 & 3.0 \\
    EchoShot~\cite{wang2025echoshot} & 8.2 & 7.0 & 5.2 & 5.4 &
    OVI~\cite{low2025ovi} & 7.8 & 7.6 & 4.6 & 4.4 \\
    CineTrans~\cite{wu2025cinetrans} & 9.4 & 6.6 & 5.0 & 5.2 &
    LTX-2~\cite{hacohen2026ltx2} & 6.6 & 7.2 & 2.4 & 2.4 \\
    IC-LoRA~\cite{huang2024iclora} + Wan~\cite{wan2025wan} & 6.6 & 4.8 & 2.8 & 3.2 &
    \textbf{Ours (MAVIN)} & \textbf{36.8} & \textbf{42.2} & \textbf{66.4} & \textbf{64.6} \\
    StoryDiff.~\cite{zhou2024storydiffusion} + Wan~\cite{wan2025wan} & 6.2 & 5.0 & 3.0 & 3.4 &
    & & & & \\
    \bottomrule
  \end{tabular}%
  }
\end{table}

\subsection{Applications}

MAVIN supports several practical creative workflows for multi-shot audio-visual generation. Editable dialogue intervals in the shot-level script allow direct control over speaking pace. The generated storyboard can also be further modified for narrative editing, allowing story progression to be adjusted without redesigning the entire prompt. Moreover, assigning visual appearances and vocal timbres through image and audio references enables flexible identity customization. Finally, the multi-agent scripting pipeline can transform free-form user prompts into structured multi-shot audio-visual narratives, simplifying script planning for professional filmmaking workflows.

\section{Conclusion}

In this paper, we present MAVIN, the first framework for multi-shot audio-visual generation with customized narrative control. To resolve the challenge of temporal misalignment across complex timelines, we introduce boundary-aware attention, which enforces temporal alignment for shot transitions, dialogue intervals, and narrative events. Furthermore, we propose ID-aware propagation to overcome the limitations of multi-subject controllability, maintaining precise character consistency via image and audio references. 
To tackle incomplete scripting, we design a multi-agent scripting pipeline that translates free-form user inputs into hierarchical captions. Finally, we construct the MAVINSet dataset to facilitate robust training and evaluation. We believe our framework opens up a new avenue for integrating generative models into professional filmmaking workflows.

\noindent \textbf{Limitation.} 
Our model can produce videos of up to 15 seconds with a maximum of 3 customized characters. In its current state, deploying our framework for practical filmmaking workflows still requires iterative prompting and manual post-processing. 
For future exploration, by integrating stronger foundation models and scaling up the parameters, its practical utility could be further improved.

\section*{Acknowledgements}
This work is supported by Beijing Major Science and Technology Project (Grant No. Z251100008125009) and National Natural Science Foundation of China (Grant No. 62136001).

\bibliographystyle{splncs04}
\bibliography{main}

\ifdefined\ARXIVCOMBINED
\clearpage
\setcounter{section}{0}
\setcounter{figure}{0}
\setcounter{table}{0}
\setcounter{equation}{0}
\renewcommand{\thesection}{\Alph{section}}
\renewcommand{\thesubfigure}{(\alph{subfigure})}
\renewcommand{\theequation}{S\arabic{equation}}
\makeatletter
\renewcommand{\thefigure}{S\@arabic\c@figure}
\renewcommand{\thetable}{S\@arabic\c@table}
\makeatother
\begingroup
\let\originalinstitute\institute
\renewcommand{\institute}[1]{%
    \originalinstitute{%
        Peking University \and
        Kling Team, Kuaishou Technology \and
        Institute of Automation, Chinese Academy of Sciences \and
        Sun Yat-sen University
        \\
        \email{liukq04@gmail.com, \{shuchenweng, shiboxin\}@pku.edu.cn}%
    }%
}
\let\originalfootnotetext\footnotetext
\renewcommand{\footnotetext}[2][]{%
    \begingroup
    \renewcommand{\thefootnote}{}
    \originalfootnotetext{$^{*}$ Equal contribution. $^{\dagger}$ Corresponding authors.}%
    \addtocounter{footnote}{-1}
    \endgroup
}
\renewcommand{\bibliographystyle}[1]{}
\renewcommand{\bibliography}[1]{}
\documentclass[runningheads]{llncs}

\usepackage{eccv}

\usepackage{eccvabbrv}
\usepackage{graphicx}
\usepackage{booktabs}
\usepackage{multirow}
\usepackage{amsmath}
\usepackage{amssymb}
\usepackage[accsupp]{axessibility}
\usepackage[pagebackref]{hyperref}
\usepackage[capitalize]{cleveref} 
\usepackage{orcidlink}

\renewcommand\thesection{\Alph{section}}

\renewcommand\thesubfigure{(\alph{subfigure})}
\renewcommand{\theequation}{S\arabic{equation}}

\makeatletter 
\renewcommand{\thefigure}{S\@arabic\c@figure}
\renewcommand{\thetable}{S\@arabic\c@table}

\makeatother

\newcommand{\todo}[1]{\textcolor{red}{[TODO: #1]}}

\begin{document}

\newcommand{\myPaperTitle}{Supplementary Material of \\ MAVIN: Multi-Shot Audio-Visual Generation with Customized Narrative Control}
\title{\myPaperTitle}
\titlerunning{Supplementary Material for MAVIN}

\author{
Kaiqi Liu\inst{1,2}$^{*}$ \and
Yunyao Mao\inst{2}$^{*}$ \and
Ziqi Cai\inst{1} \and
Zheng Geng\inst{3} \and
Jing Wang\inst{4} \and\\[-0.1em]
Qiulin Wang\inst{2} \and
Xintao Wang\inst{2} \and
Pengfei Wan\inst{2} \and
Kun Gai\inst{2} \and\\[-0.1em]
Shuchen Weng\inst{1}$^{\dagger}$ \and
Boxin Shi\inst{1}$^{\dagger}$
}

\authorrunning{K.~Liu et al.}

\institute{
Peking University, Beijing, China \and
Kling Team, Kuaishou Technology, Beijing, China \and
Institute of Automation, Chinese Academy of Sciences, Beijing, China \and
Sun Yat-sen University, Shenzhen, China
\\
\email{liukq04@gmail.com, \{shuchenweng, shiboxin\}@pku.edu.cn}
}
\maketitle

\footnotetext[1]{Kaiqi Liu is with School of Software and Microelectronics, Peking University. 
Ziqi Cai, Shuchen Weng, and Boxin Shi are with State Key Laboratory of Multimedia Information Processing and National Engineering Research Center of Visual Technology, School of Computer Science, Peking University.
\par\noindent $^{*}$ Equal contribution. 
$^{\dagger}$ Corresponding authors.}

\section{Applications}

\begin{figure}[t]
\centerline{
\includegraphics[width=1.0\columnwidth]{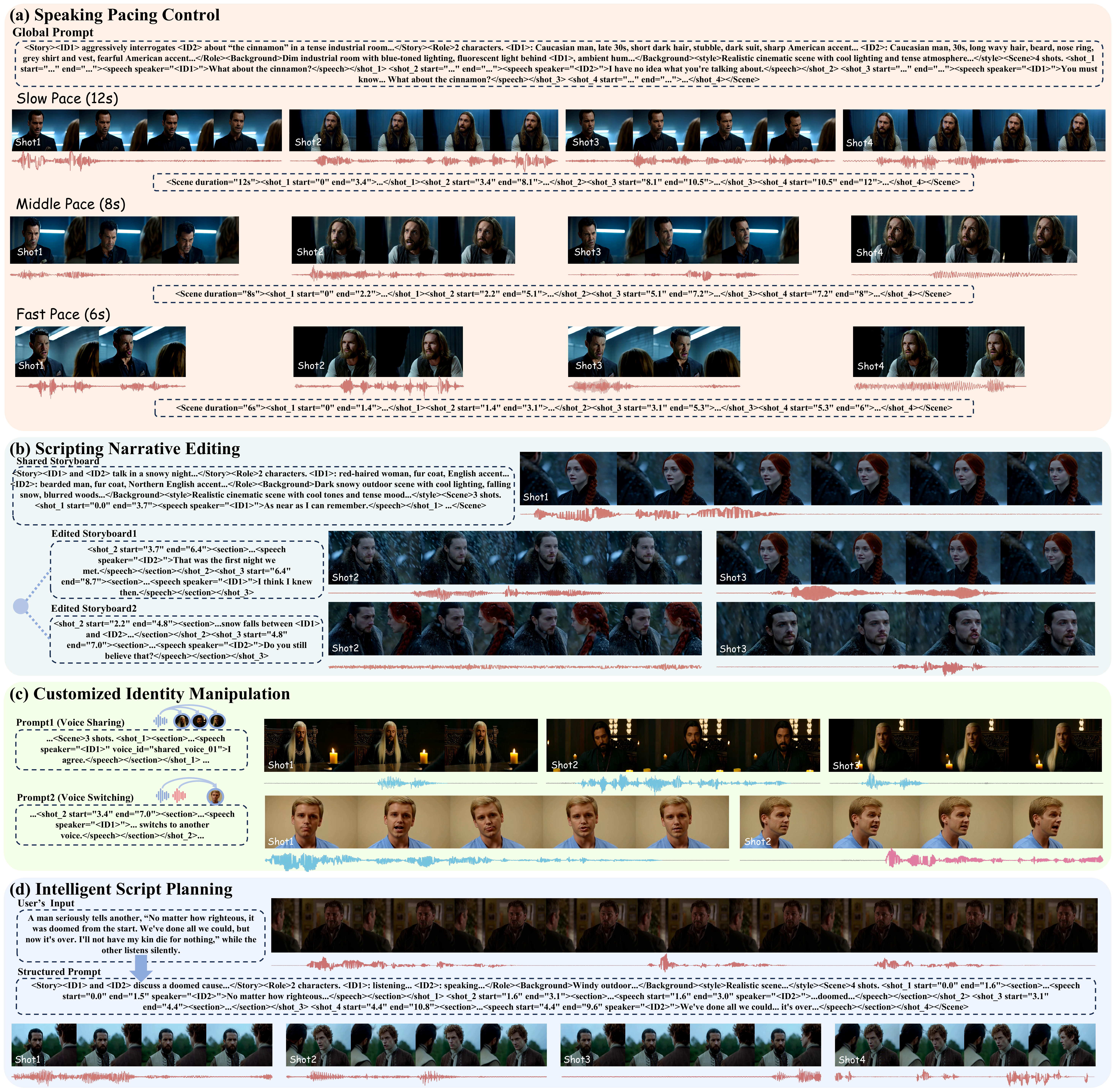}
}
\caption{
Representative application scenarios. Our proposed MAVIN framework enables highly practical and controllable creative workflows for professional filmmaking.
}

\label{fig:supp_application}
\end{figure}

Beyond standard audio-visual generation, our framework supports several highly practical creative workflows.
We present representative application scenarios in \cref{fig:supp_application}, details as follows. Please refer to the supplementary video for comprehensive results.

\noindent\textbf{Speaking pacing control.}
Our framework enables explicitly specifying dialogue intervals by providing shot-level prompts (\eg, ``ID\_A replies 1.9--2.5s: `You knew--'''). As presented in~\cref{fig:supp_application}~(a), this makes it possible for users to intuitively adjust the speaking pace simply by shortening or prolonging these dialogue intervals.

\noindent\textbf{Scripting narrative editing.}
Since the narrative script is text-based and directly accessible to users, once the script is generated by the multi-agent scripting pipeline, users can flexibly modify it according to their creative intentions (\eg, steering the story's progression in a different direction). As presented in~\cref{fig:supp_application}~(b), this effectively lowers the barrier to professional filmmaking.

\noindent\textbf{Customized identity manipulation.}
With the proposed ID-aware propagation, users can provide image and audio references, freely assigning specific visual appearances and vocal timbres to characters within the cinematic narrative (\eg, assigning the same character different vocal styles across scenes). As presented in~\cref{fig:supp_application}~(c), this makes the generation process highly flexible and creative.

\noindent\textbf{Intelligent script planning.}
Our multi-agent scripting pipeline effectively transforms users' free-form prompts (\eg, a single sentence) into comprehensive audio-visual narratives (\ie, hierarchical captions with decoupled global, shot, and role-level narrative semantics). As presented in~\cref{fig:supp_application}~(d), this significantly simplifies script planning for professional filmmaking workflows.

\section{Additional Experiments}
We provide additional experiments and analyses to further validate the effectiveness, robustness, and generalization ability of MAVIN.

\subsection{Design vs. Annotation Gains}

To disentangle architectural contributions from the gains brought by high-quality data and annotations, we conduct a controlled fine-tuning experiment based on the OVI backbone. Specifically, we fine-tune OVI on our curated training set while removing all MAVIN-specific designs, denoted as OVI-FT. 
As shown in~\cref{tab:design_annotation_gains}, OVI-FT improves upon the original OVI backbone on most metrics, confirming the benefit of our curated multi-shot audio-visual dataset. Nevertheless, MAVIN consistently outperforms OVI-FT across all metrics. These results demonstrate that the performance gains of MAVIN cannot be attributed to annotation quality alone. Instead, the proposed boundary-aware attention and ID-aware propagation provide essential architectural benefits for temporally aligned and identity-consistent multi-shot audio-visual generation.

\begin{table*}[!t]
\centering
\small
\setlength{\tabcolsep}{3.2pt}
\caption{Controlled analysis for disentangling architectural gains from data and annotation gains. OVI-FT denotes the OVI backbone fine-tuned on our curated training set without MAVIN-specific designs.}
\label{tab:design_annotation_gains}
\resizebox{\textwidth}{!}{
\begin{tabular}{lccccccccccccc}
\toprule
\multirow{2}{*}{Method} &
\multicolumn{2}{c}{Quality} &
\multicolumn{3}{c}{Semantics} &
\multicolumn{3}{c}{AV-Alignment} &
\multicolumn{5}{c}{Multi-shot Consistency} \\
\cmidrule(lr){2-3}
\cmidrule(lr){4-6}
\cmidrule(lr){7-9}
\cmidrule(lr){10-14}
& FVD $\downarrow$ & FAD $\downarrow$
& TVS $\uparrow$ & TAS $\uparrow$ & WER $\downarrow$
& Sync $\uparrow$ & AV-IB $\uparrow$ & TAMS $\uparrow$
& SC $\uparrow$ & BC $\uparrow$ & CISC $\uparrow$ & BISC $\uparrow$ & STA $\uparrow$ \\
\midrule

OVI~\cite{low2025ovi}
& 318.4 & 8.3
& 0.1983 & 0.2203 & 0.093
& 4.231 & 0.227 & 0.4432
& 0.9565 & 0.9578 & 0.6121 & 0.7789 & 0.4992 \\

OVI-FT
& 272.4 & 7.9
& 0.2187 & 0.2193 & 0.069
& 5.012 & 0.236 & 0.5371
& 0.9596 & 0.9587 & 0.6127 & 0.7796 & 0.5442 \\

Ours (MAVIN)
& \textbf{231.6} & \textbf{6.8}
& \textbf{0.2471} & \textbf{0.2392} & \textbf{0.048}
& \textbf{6.032} & \textbf{0.263} & \textbf{0.8104}
& \textbf{0.9695} & \textbf{0.9657} & \textbf{0.6319} & \textbf{0.8013} & \textbf{0.9897} \\

\bottomrule
\end{tabular}
}
\end{table*}

\subsection{OOD Generalization}

To evaluate the broader generalization ability of MAVIN beyond MAVINSet, we construct two additional out-of-distribution evaluation benchmarks. First, we curate an OOD benchmark from MSVBench~\cite{shi2026msvbench}, which contains domains and scenarios different from MAVINSet. Second, we build a real-world benchmark (RWBench), consisting of 240 newly collected YouTube videos across 15 categories with lighter curation. Both benchmarks are adapted into multi-shot audio-visual format captions for evaluation.
As shown in~\cref{tab:ood_generalization}, MAVIN consistently outperforms the strongest cascaded baseline CineTrans and the strongest joint baseline LTX-2 on both benchmarks. On MSVBench, MAVIN achieves substantial improvements in temporal alignment and shot-transition control, improving TAMS from 0.5427 to 0.7923 and STA from 0.5896 to 0.9775 over the best baseline. Similar trends are observed on RWBench, where MAVIN achieves the best performance across all metrics. These results demonstrate that the advantages of MAVIN generalize to diverse real-world scenarios and are not limited to the MAVINSet distribution.

\begin{table*}[!t]
\centering
\small
\setlength{\tabcolsep}{3.6pt}
\caption{OOD generalization results on MSVBench and our less-curated Real-World Benchmark (RWBench). MAVIN consistently outperforms the strongest cascaded and joint baselines across diverse real-world scenarios.}
\label{tab:ood_generalization}
\resizebox{\textwidth}{!}{
\begin{tabular}{lccccccccccc}
\toprule
Method
& TVS $\uparrow$ & TAS $\uparrow$ & WER $\downarrow$
& Sync $\uparrow$ & AV-IB $\uparrow$ & TAMS $\uparrow$
& SC $\uparrow$ & BC $\uparrow$ & CISC $\uparrow$ & BISC $\uparrow$ & STA $\uparrow$ \\
\midrule

\multicolumn{12}{l}{\textit{- OOD Benchmark: MSVBench}} \\
CineTrans~\cite{wu2025cinetrans}
& 0.2068 & 0.2036 & --
& 1.846 & 0.196 & --
& 0.9398 & 0.9571 & 0.6124 & 0.7526 & 0.5896 \\

LTX-2~\cite{hacohen2026ltx2}
& 0.1972 & 0.2315 & 0.091
& 4.512 & 0.199 & 0.5427
& 0.9612 & 0.9517 & 0.5963 & 0.7861 & 0.5648 \\

Ours (MAVIN)
& \textbf{0.2513} & \textbf{0.2361} & \textbf{0.052}
& \textbf{5.921} & \textbf{0.259} & \textbf{0.7923}
& \textbf{0.9721} & \textbf{0.9682} & \textbf{0.6385} & \textbf{0.8074} & \textbf{0.9775} \\

\midrule
\multicolumn{12}{l}{\textit{- Less-Curated Real-World Benchmark: RWBench}} \\
CineTrans~\cite{wu2025cinetrans}
& 0.1995 & 0.2094 & --
& 1.721 & 0.203 & --
& 0.9356 & 0.9602 & 0.5961 & 0.7614 & 0.5486 \\

LTX-2~\cite{hacohen2026ltx2}
& 0.2041 & 0.2248 & 0.118
& 3.982 & 0.207 & 0.4928
& 0.9487 & 0.9563 & 0.5903 & 0.7925 & 0.5419 \\

Ours (MAVIN)
& \textbf{0.2447} & \textbf{0.2415} & \textbf{0.061}
& \textbf{5.603} & \textbf{0.265} & \textbf{0.7468}
& \textbf{0.9584} & \textbf{0.9701} & \textbf{0.6172} & \textbf{0.8126} & \textbf{0.9764} \\

\bottomrule
\end{tabular}
}
\end{table*}

\subsection{Robustness to Imperfect Scripts}

\begin{figure*}[!t]
\centering
\includegraphics[width=\textwidth]{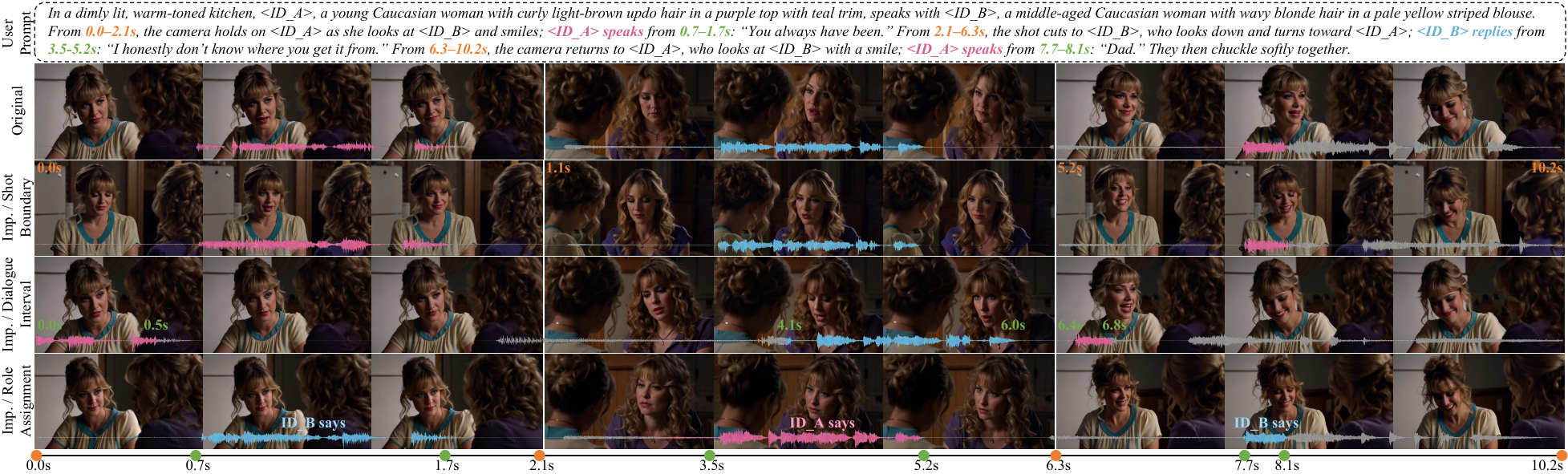}
\caption{Robustness to imperfect scripts. MAVIN maintains coherent audio-visual generation under perturbed shot boundaries, dialogue intervals, and role assignments.}
\label{fig:supp_imperfect_scripts}
\end{figure*}

In practical creative workflows, user-provided or automatically generated scripts may contain imperfect temporal or role-level specifications. To evaluate the robustness of MAVIN under such imperfect conditions, we randomly perturb shot boundaries, dialogue intervals, and role assignments in the hierarchical captions. These perturbations simulate common script-level errors, including inaccurate shot timing, shifted speaking intervals, and ambiguous speaker-role bindings.
As shown in~\cref{fig:supp_imperfect_scripts}, MAVIN remains robust under these perturbed conditions. When shot boundaries are shifted, the model still preserves coherent visual progression across shots. When dialogue intervals are perturbed, the generated speech remains temporally localized and synchronized with the corresponding visual events. When role assignments are perturbed, MAVIN still maintains reasonable identity consistency without severe cross-role contamination. These results suggest that MAVIN can tolerate moderate script imperfections and remains applicable to realistic scripting workflows.

\subsection{Generalization to More Characters}

\begin{figure}[!t]
\centering
\includegraphics[width=\linewidth]{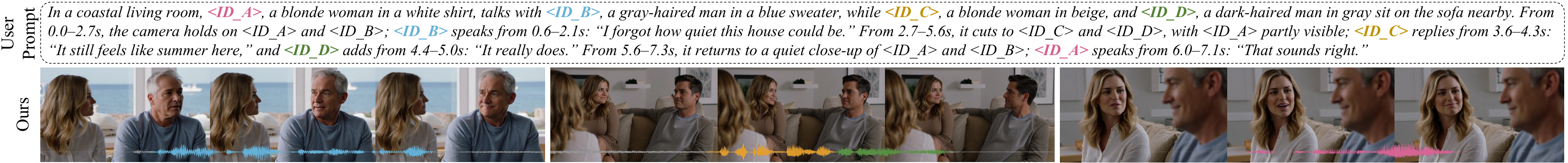}
\caption{Generalization to more characters. After fine-tuning on curated four-character data, MAVIN can generate coherent multi-shot audio-visual results with four characters.}
\label{fig:supp_four_characters}
\end{figure}

As clean multi-shot audio-visual clips with more characters are relatively scarce, MAVIN adopts a maximum three-character setting in the main experiments to ensure data quality. To further evaluate its potential for more complex multi-character scenarios, we curate additional four-character data and fine-tune MAVIN under the same framework.
As shown in~\cref{fig:supp_four_characters}, MAVIN can generate coherent four-character audio-visual results while maintaining reasonable identity consistency and dialogue synchronization. 
These results suggest that the proposed framework has the potential to extend beyond the three-character setting, and its multi-character capability can be further improved with more high-quality training data.

\subsection{Additional Comparison with Cascaded Models}
We further extend the quantitative evaluation with cascaded Text-to-Audio (T2A) and Audio-to-Video (A2V) pipelines, where an audio track is first generated and subsequently used to drive video generation models. The quantitative results are presented in \cref{tab:standardcomparison}. 
We also include additional Text-to-Video (T2V) and Video-to-Audio (V2A) cascaded baselines, with results presented in \cref{tab:expanded_cascaded}.

\noindent \textbf{Metrics.}
We conduct this evaluation using the same metrics detailed in Sec.~5.2 of the main paper, including quality (FVD and FAD), semantics (TVS, TAS, and WER), AV-alignment (Sync, AV-IB, and TAMS), and multi-shot consistency (SC, BC, CISC, BISC, and STA).

\noindent \textbf{Baselines.}
Following common practice in prior audio-visual generation works~\cite{liu2026javisditplusplus, wang2026apollo, qiang2026mmsonate}, we construct representative cascaded baselines by separately pairing AudioLDM2~\cite{liu2023audioldm} with TempoTokens~\cite{yariv2024tempotokens} and MTV~\cite{weng2025mtv}.
To further expand the cascaded comparison, we evaluate CineTrans~\cite{wu2025cinetrans} with See\&Hear~\cite{xing2024seeing} and MMAudio~\cite{cheng2025mmaudio} as additional V2A backbones, and include StoryMem~\cite{zhang2025storymem} with Hunyuan-Foley~\cite{shan2025hunyuanvideofoley} as an additional multi-shot T2V-based cascaded baseline.

\noindent \textbf{Analysis.}
For visual quality and semantic consistency, MAVIN achieves stronger FVD and TVS scores than the evaluated cascaded baselines. Furthermore, since current T2A models (\eg, AudioLDM2~\cite{liu2023audioldm}) struggle to generate intelligible speech, the WER and TAMS metrics are inapplicable to these baselines. Moreover, MAVIN achieves significantly stronger lip synchronization (Sync).
As shown in~\cref{tab:expanded_cascaded}, MAVIN also outperforms the expanded T2V+V2A cascaded baselines. Although stronger cascaded components improve certain quality or semantic metrics, the decoupled generation process still limits joint audio-visual modeling and temporal controllability across shot boundaries, leading to inferior Sync, AV-IB, and STA scores.

\begin{table*}[!t]
\centering
\small
\setlength{\tabcolsep}{3.2pt}
\caption{Additional quantitative comparison with cascaded T2A and A2V baselines. The best results are highlighted with bold, and `-' indicates the metric is inapplicable.}
\label{tab:standardcomparison}
\resizebox{\textwidth}{!}{
\begin{tabular}{l|cc|ccc|ccc|ccccc}
\toprule
\multirow{2}{*}{Method} &
\multicolumn{2}{c|}{Quality} &
\multicolumn{3}{c|}{Semantics} &
\multicolumn{3}{c|}{AV-Alignment} &
\multicolumn{5}{c}{Multi-shot Consistency} \\
\cmidrule(lr){2-3}
\cmidrule(lr){4-6}
\cmidrule(lr){7-9}
\cmidrule(lr){10-14}
& FVD $\downarrow$ & FAD $\downarrow$
& TVS $\uparrow$ & TAS $\uparrow$ & WER $\downarrow$
& Sync $\uparrow$ & AV-IB $\uparrow$ & TAMS $\uparrow$
& SC $\uparrow$ & BC $\uparrow$ & CISC $\uparrow$ & BISC $\uparrow$ & STA $\uparrow$ \\
\midrule

\multicolumn{14}{l}{\textit{- Cascaded Generation Models: T2A+A2V}} \\

TempoTokens~\cite{yariv2024tempotokens}
& 802.4 & 17.3
& 0.097 & 0.1793 & -
& 2.35 & 0.117 & -
& 0.9601 & 0.9621 & 0.3137 & 0.4005 & - \\

MTV~\cite{weng2025mtv}
& 382.9 & 17.3
& 0.185 & 0.1793 & -
& 3.890 & 0.220 & -
& 0.9592 & 0.9586 & 0.5862 & 0.7325 & - \\

\midrule
\multicolumn{14}{l}{\textit{- Joint Generation Models: T2AV}} \\

Ours (MAVIN)
& \textbf{231.6} & \textbf{6.8}
& \textbf{0.2471} & \textbf{0.2392} & \textbf{0.048}
& \textbf{6.032} & \textbf{0.263} & \textbf{0.8104}
& \textbf{0.9695} & \textbf{0.9657} & \textbf{0.6319} & \textbf{0.8013} & \textbf{0.9897} \\

\bottomrule
\end{tabular}
}
\end{table*}

\begin{table*}[!t]
\centering
\small
\setlength{\tabcolsep}{3.6pt}
\caption{Additional quantitative comparison with expanded T2V+V2A cascaded baselines. The best results are highlighted with bold.}
\label{tab:expanded_cascaded}
\resizebox{\textwidth}{!}{
\begin{tabular}{l|cc|cc|cc|ccccc}
\toprule
\multirow{2}{*}{Method} &
\multicolumn{2}{c|}{Quality} &
\multicolumn{2}{c|}{Semantics} &
\multicolumn{2}{c|}{AV-Alignment} &
\multicolumn{5}{c}{Multi-shot Consistency} \\
\cmidrule(lr){2-3}
\cmidrule(lr){4-5}
\cmidrule(lr){6-7}
\cmidrule(lr){8-12}
& FVD $\downarrow$ & FAD $\downarrow$
& TVS $\uparrow$ & TAS $\uparrow$
& Sync $\uparrow$ & AV-IB $\uparrow$
& SC $\uparrow$ & BC $\uparrow$ & CISC $\uparrow$ & BISC $\uparrow$ & STA $\uparrow$ \\
\midrule

\multicolumn{12}{l}{\textit{- Cascaded Generation Models: Additional V2A Backbones}} \\

See\&Hear~\cite{xing2024seeing}
& 254.2 & 16.9
& 0.2013 & 0.2073
& 2.103 & 0.199
& 0.9425 & 0.9543 & 0.6073 & 0.7603 & 0.5820 \\

MMAudio~\cite{cheng2025mmaudio}
& 254.2 & 12.8
& 0.2013 & 0.2198
& 2.227 & 0.213
& 0.9425 & 0.9543 & 0.6073 & 0.7603 & 0.5820 \\

\midrule
\multicolumn{12}{l}{\textit{- Cascaded Generation Models: Additional T2V Backbones}} \\

StoryMem~\cite{zhang2025storymem}
& 259.3 & 12.1
& 0.2163 & 0.2124
& 2.113 & 0.212
& 0.9416 & 0.9502 & 0.6216 & 0.8007 & 0.5606 \\

\midrule
\multicolumn{12}{l}{\textit{- Joint Generation Models: T2AV}} \\

Ours (MAVIN)
& \textbf{231.6} & \textbf{6.8}
& \textbf{0.2471} & \textbf{0.2392}
& \textbf{6.032} & \textbf{0.263}
& \textbf{0.9695} & \textbf{0.9657} & \textbf{0.6319} & \textbf{0.8013} & \textbf{0.9897} \\

\bottomrule
\end{tabular}
}
\end{table*}

\subsection{Additional Comparison with Personalized Methods}
\begin{figure}[t]
\centerline{
\includegraphics[width=1.0\columnwidth]{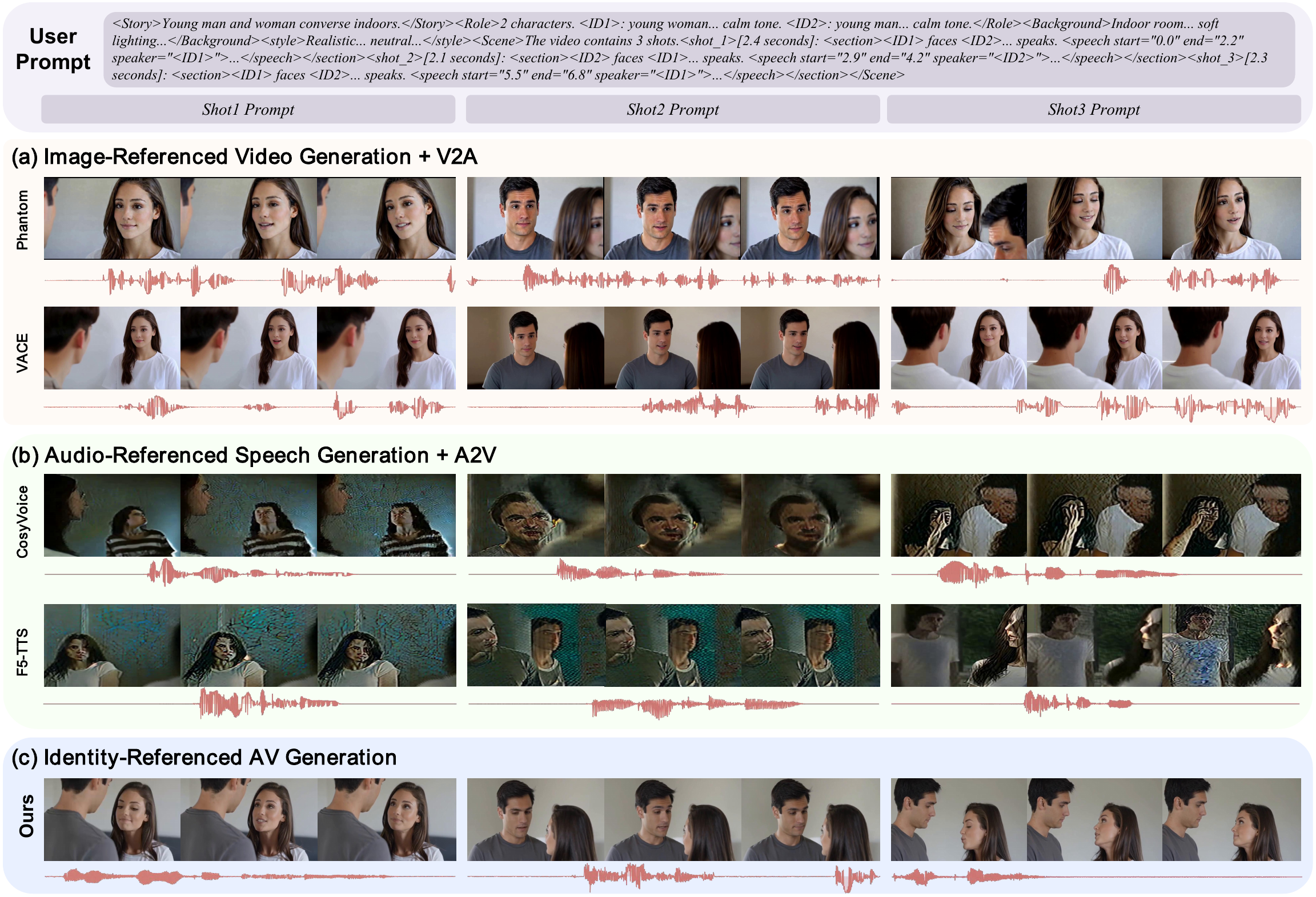}
}
\caption{
Qualitative comparison with personalized methods. Given the same image and audio references, MAVIN achieves superior identity preservation.
}

\label{fig:supp_personalized}
\end{figure}

We additionally compare our MAVIN with personalized generation baselines to evaluate its capability for identity customization using reference images and audio. The qualitative and quantitative results are presented in \cref{fig:supp_personalized} and \cref{tab:personalized_sup}, respectively.

\noindent\textbf{Metrics.}
For personalized evaluation, we use ArcFace~\cite{deng2019arcface} to extract visual features from the reference portrait images and the generated face crops, calculating the cosine similarity to measure Visual Consistency (VC).
Similarly, we use WavLM~\cite{chen2022wavlm} to extract audio features from the reference vocal track and the generated speech segments of the same identity, calculating the cosine similarity to measure Audio Consistency (AC).

\noindent\textbf{Baselines.}
To enable a fair comparison, we complement the audio and video modalities for each baseline method.
Specifically, image-referenced video generation methods~\cite{liu2025phantom,jiang2025vace} are paired with a V2A model (\ie, Hunyuan-Foley~\cite{shan2025hunyuanvideofoley}) for downstream audio generation. 
Conversely, audio-referenced speech generation methods~\cite{du2024cosyvoice,chen2025f5tts} are paired with a A2V model (\ie, MTV~\cite{weng2025mtv}) for downstream video generation.

\noindent \textbf{Analysis.}
As shown in~\cref{tab:personalized_sup}, image-referenced video generation methods~\cite{liu2025phantom,jiang2025vace} typically lack fine-grained lip synchronization, causing downstream V2A models to fail at generating intelligible speech. Consequently, the WER and TAMS metrics are inapplicable for these methods.
On the other hand, while audio-referenced speech generation methods~\cite{du2024cosyvoice,chen2025f5tts} provide customized vocals, they typically ignore ambient sound effects. Furthermore, the limited capabilities of downstream A2V models degrade the overall visual quality, resulting in inferior AV-IB and FVD scores.

\subsection{Pairwise Human Evaluation}

To further validate human preference under a pairwise comparison protocol, we conduct an additional A/B user study. In each comparison, raters are presented with the results from MAVIN and one baseline under randomized order and hidden method names. They are asked to select the preferred result according to four dimensions: Audio-Visual Quality (AVQ), Multi-shot Identity Consistency (MIC), Audio Narrative Adherence (ANA), and Video Narrative Adherence (VNA). We report the win-rate of MAVIN against each baseline in \cref{tab:pairwise_human_eval}. 
MAVIN achieves consistent preference gains over both cascaded and joint generation baselines across all evaluation dimensions. These results further support MAVIN's perceptual quality, identity consistency, and narrative controllability under pairwise human evaluation.

\begin{table*}[!t]
\centering
\caption{Pairwise human evaluation results. We report MAVIN's win-rate (\%) against each baseline under randomized order and hidden method names.}
\label{tab:pairwise_human_eval}
\resizebox{\textwidth}{!}{
\begin{tabular}{lcccc lcccc}
\toprule
Baseline & AVQ $\uparrow$ & MIC $\uparrow$ & ANA $\uparrow$ & VNA $\uparrow$
& Baseline & AVQ $\uparrow$ & MIC $\uparrow$ & ANA $\uparrow$ & VNA $\uparrow$ \\
\midrule

VideoGen-of-Thought~\cite{zheng2024videogenofthought}
& 78.6 & 76.8 & 86.9 & 86.1
& JavisDiT~\cite{liu2025javisdit}
& 82.7 & 83.9 & 89.1 & 88.4 \\

MovieAgent~\cite{wu2025automatedmovieagent}
& 81.2 & 82.6 & 88.1 & 87.4
& UniVerse-1~\cite{wang2025universe1}
& 79.8 & 81.5 & 87.3 & 86.8 \\

EchoShot~\cite{wang2025echoshot}
& 71.4 & 74.2 & 81.3 & 80.5
& OVI~\cite{low2025ovi}
& 66.8 & 68.1 & 76.4 & 75.9 \\

CineTrans~\cite{wu2025cinetrans}
& 65.7 & 75.1 & 77.2 & 76.4
& LTX-2~\cite{hacohen2026ltx2}
& 69.2 & 68.9 & 82.6 & 81.7 \\

IC-LoRA~\cite{huang2024iclora} + Wan~\cite{wan2025wan}
& 75.3 & 80.7 & 87.6 & 86.2
& See\&Hear~\cite{xing2024seeing}
& 69.4 & 75.1 & 82.7 & 76.4 \\

StoryDiff.~\cite{zhou2024storydiffusion} + Wan~\cite{wan2025wan}
& 76.1 & 79.5 & 87.1 & 85.8
& MMAudio~\cite{cheng2025mmaudio}
& 68.5 & 75.1 & 72.6 & 76.4 \\

StoryMem~\cite{zhang2025storymem}
& 63.8 & 74.5 & 78.3 & 76.9
&  &  &  &  &  \\

\bottomrule
\end{tabular}
}
\end{table*}

\begin{table*}[t]
\centering
\small
\setlength{\tabcolsep}{3.2pt}
\caption{Additional quantitative comparison with personalized methods.}
\label{tab:personalized_sup}
\resizebox{\textwidth}{!}{
\begin{tabular}{lccccccccccccccc}
\toprule
\multirow{2}{*}{Method} &
\multicolumn{2}{c}{Quality} &
\multicolumn{3}{c}{Semantics} &
\multicolumn{3}{c}{AV-Alignment} &
\multicolumn{5}{c}{Multi-shot Consistency} &
\multicolumn{2}{c}{Personalization} \\
\cmidrule(lr){2-3}
\cmidrule(lr){4-6}
\cmidrule(lr){7-9}
\cmidrule(lr){10-14}
\cmidrule(lr){15-16}
& FVD $\downarrow$ & FAD $\downarrow$
& TVS $\uparrow$ & TAS $\uparrow$ & WER $\downarrow$
& Sync $\uparrow$ & AV-IB $\uparrow$ & TAMS $\uparrow$
& SC $\uparrow$ & BC $\uparrow$ & CISC $\uparrow$ & BISC $\uparrow$ & STA $\uparrow$
& VC $\uparrow$ & AC $\uparrow$ \\
\midrule

\multicolumn{16}{l}{\textit{- Cascaded: Image-Referenced Video Generation + V2A}} \\
Phantom~\cite{liu2025phantom}
& 275.3 & 11.7
& \underline{0.2318} & 0.2189 & --
& 2.396 & 0.202 & --
& 0.9663 & 0.9546 & \underline{0.6026} & 0.7149 & --
& \textbf{0.6285} & -- \\

VACE~\cite{jiang2025vace}
& \underline{253.4} & 12.3
& 0.2296 & 0.2293 & --
& 2.513 & 0.212 & --
& \underline{0.9677} & 0.9574 & 0.5834 & 0.6997 & --
& 0.6049 & -- \\

\midrule
\multicolumn{16}{l}{\textit{- Cascaded: Audio-Referenced Speech Generation + A2V}} \\
CosyVoice~\cite{du2024cosyvoice}
& 324.2 & \underline{7.2}
& 0.2014 & 0.2205 & 0.053
& \underline{3.821} & \underline{0.235} & --
& 0.9583 & 0.9584 & 0.5739 & 0.7572 & --
& -- & \underline{0.5002} \\

F5-TTS~\cite{chen2025f5tts}
& 336.4 & 7.5
& 0.2099 & 0.2186 & \textbf{0.046}
& 3.804 & 0.229 & --
& 0.9591 & \underline{0.9589} & 0.5813 & \underline{0.7669} & --
& -- & \textbf{0.5610} \\

\midrule
\multicolumn{16}{l}{\textit{- Joint: Identity-Referenced AV Generation}} \\
Ours (MAVIN)
& \textbf{241.9} & \textbf{6.9}
& \textbf{0.2463} & \textbf{0.2391} & \underline{0.049}
& \textbf{6.057} & \textbf{0.266} & \textbf{0.793}
& \textbf{0.9682} & \textbf{0.9653} & \textbf{0.6397} & \textbf{0.7996} & \textbf{0.9886}
& \underline{0.6259} & 0.4716 \\
\bottomrule
\end{tabular}
}
\end{table*}

\begin{figure}[t]
\centerline{
\includegraphics[width=1.0\columnwidth]{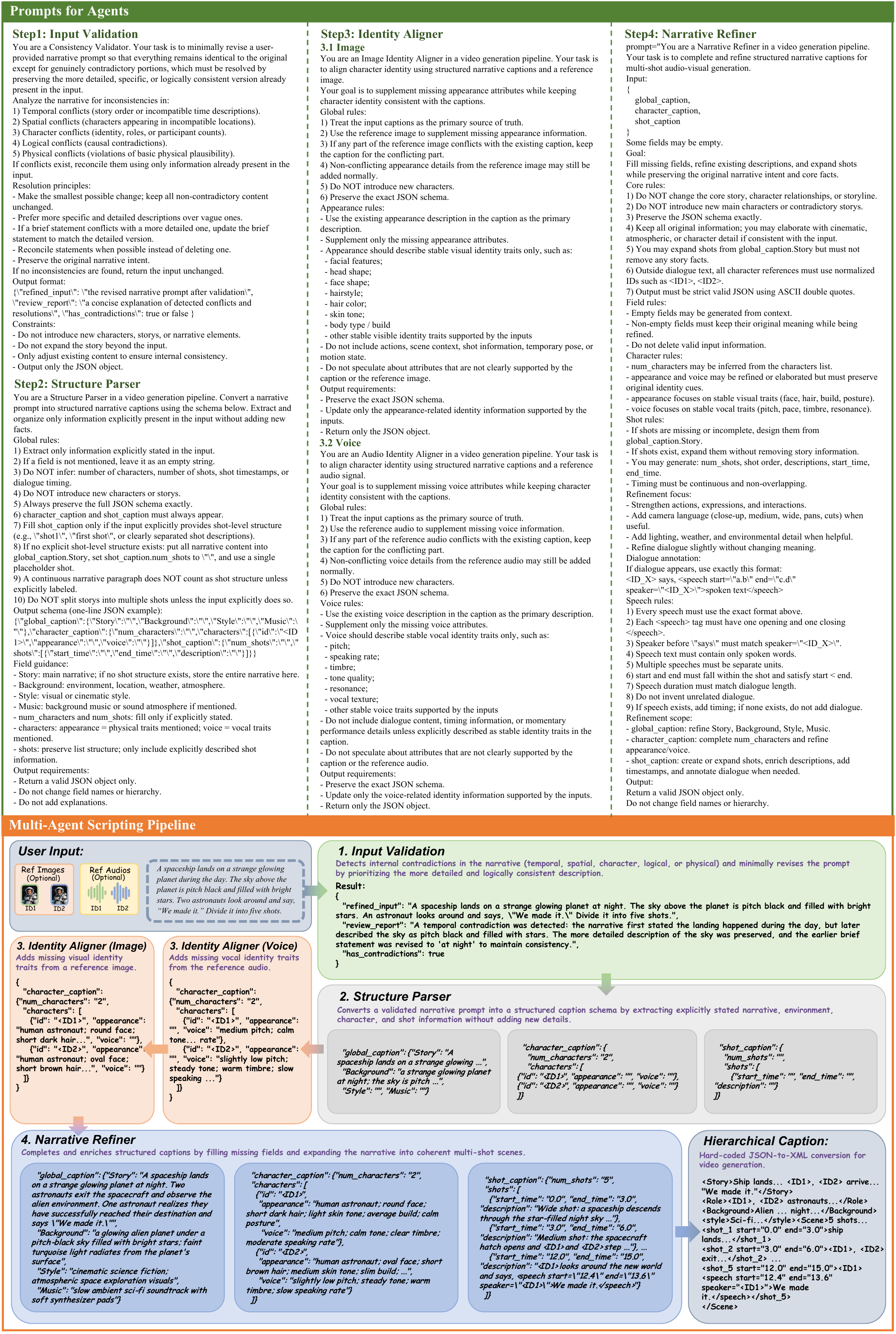}
}
\caption{
An example of our multi-agent scripting pipeline.
}

\label{fig:supp_agent}
\end{figure}

\section{Additional Details}

To ensure reproducible results and a transparent workflow, we provide comprehensive details for the pipeline, metrics, and dataset.

\subsection{Multi-Agent Scripting Pipeline Details}
As introduced in Sec.~4.4 of the main paper, we employ a multi-agent scripting pipeline to transform free-form user prompts into hierarchical captions. A detailed example is illustrated in \cref{fig:supp_agent}, and the specific components are described as follows.

\noindent\textbf{Input validation.}
We first perform an initial input validation process. We utilize a VLM~\cite{xu2025qwen3omni} to detect temporal, spatial, logical, and role-level conflicts (\eg, inconsistent shot order, contradictory actions, and ambiguous speaker assignments). The model attempts to resolve these conflicts by reorganizing the user input, while being strictly prohibited from introducing any information not provided by the user. If the conflicts persist, the input is considered invalid, and the process is terminated.

\noindent\textbf{Structure parser.}
The structure parser converts the validated prompt into a temporally grounded shot plan. It decomposes the narrative into cinematic shots with explicit temporal boundaries $\{\mathcal{T}_i^\mathrm{s}\}_{i=1}^{S}$, identifies the role set $\mathcal{R}$, and extracts dialogue spans $\{T_j^\mathrm{r}\}_{j \in \mathcal{R}}$ for each role. This step establishes the shot structure utilized by our boundary-aware attention mechanism.

\noindent\textbf{Identity aligner.}
When the user provides optional image or audio references, the identity aligner binds these references to the corresponding characters. 
It summarizes stable appearance cues, speaking traits, and identity tags into role-level descriptions, which subsequently serve as the explicit conditioning signals for our ID-aware propagation.

\noindent\textbf{Narrative refiner.}
Given the established shot structure and role bindings, the narrative refiner enriches the script with shot-level actions, camera language, ambient audio, and dialogue phrasing, while strictly preserving the previously determined timestamps and identities. The final output is a hierarchical caption:

\begin{equation}
\mathcal{C}
=
\Big\{
C^{\mathrm{global}},
\{(C_i^{\mathrm{shot}}, T_i^s)\}_{i=1}^{S},
\{(C_j^{\mathrm{role}}, T_j^r)\}_{j \in \mathcal{R}}
\Big\}.
\end{equation}

\subsection{Additional Implementation Details}
While the primary training configurations are presented in Sec.~5.1 of the main paper, we provide further implementation details as follows.

\noindent \textbf{Optimization.}
We optimize the framework using the AdamW~\cite{loshchilov2017adamw} optimizer ($\beta_1=0.9$, $\beta_2=0.95$, $\epsilon=1\times10^{-8}$). 
In practice, the flow matching losses for the video and audio branches are assigned weights of 0.85 and 0.15, respectively. 
During progressive training, the model is trained for 60K, 20K, and 15K steps across the respective stages. 
All stages are conducted with a learning rate of $5\times10^{-5}$. The total training takes approximately 120 hours on 40 GPUs to reach convergence. 
During inference, it takes 289 seconds to generate an audio-visual result on a single GPU.

\noindent\textbf{Dynamic resolution and duration.}
To be compatible with diverse aspect ratios and clip lengths, we adopt a bucket-based dynamic sampling strategy. Each training sample is assigned to a predefined 480p resolution bucket with a specific aspect ratio and is center-cropped before VAE encoding. All bucket resolutions are strictly divisible by the model patch size to ensure compatible latent tokenization. 
Video clips have durations between 3 and 15 seconds. The frame count is adjusted to the maximum possible value that satisfies the temporal downsampling constraint of $4k+1$. Audio segments are synchronously trimmed within the same temporal window to maintain strict audio-visual alignment.

\subsection{Evaluation Metrics Details}
As illustrated in Sec.~5.2 of the main paper, we adopt comprehensive metrics for evaluating our framework. 
Among these, all standard metrics (\ie, FVD, FAD, TVS, TAS, WER, Sync, and AV-IB) follow their original implementations. Since previous works lack specific metrics for multi-shot evaluation, we additionally introduce four novel metrics. The detailed formulations are provided as follows.

\noindent\textbf{Time Alignment Metric for Speech (TAMS).}
To evaluate whether the generated speech occurs within the corresponding temporal segments, we introduce a time alignment metric based on the temporal Intersection-over-Union (IoU) between predicted speech activity and ground-truth speech intervals.
For each sample, we obtain the ground-truth speech segments from annotated timestamps $\mathcal{G} = \bigcup_{i=1}^{M} g_i$ and the predicted speech segments using a voice activity detector~\cite{webrtcvad} $\mathcal{P} = \bigcup_{j=1}^{N} p_j$, where $g_i$ and $p_j$ denote the $i$-th ground-truth and $j$-th predicted speech intervals, and $M$ and $N$ represent the total number of intervals, respectively.
TAMS is then defined as the temporal IoU between the predicted and ground-truth speech intervals:
\begin{equation}
\mathrm{TAMS} =
\frac{|\mathcal{G} \cap \mathcal{P}|}{|\mathcal{G} \cup \mathcal{P}|},
\end{equation}
where $|\cdot|$ denotes the total temporal duration of an interval set.
This metric penalizes both missing speech (false negatives) and speech occurring outside the designated intervals (false positives), thereby measuring whether the generated speech aligns precisely with the intended timestamps.

\noindent\textbf{Character and Background Inter-shot Consistency (CISC/BISC).}
To measure visual consistency across different shots, a straightforward approach is to extract ViCLIP~\cite{wang2023internvid} features for each shot and compute the global cosine similarity between them. However, different shots may intentionally depict different characters or scenes, and directly comparing all shots may ignore such diversity and lead to biased measurements.
To obtain a more reliable estimate, we group shots according to shared identities. 
Specifically, let $\mathcal{P}_{\mathrm{char}}$ and $\mathcal{P}_{\mathrm{bg}}$ denote the sets of all predicted shot pairs $(i, j)$ that are annotated to share the same character and background scene, respectively. For each shot $i$, we extract its visual feature $p_i$ using a pre-trained ViCLIP model. The final Character Inter-shot Consistency (CISC) and Background Inter-shot Consistency (BISC) are formally defined as the average cosine similarity over these matched pairs:
\begin{equation}
    \mathrm{CISC}=\frac{1}{|\mathcal{P}_{\mathrm{char}}|}\sum_{(i,j)\in\mathcal{P}_{\mathrm{char}}}\!\!\!\! \cos(p_i,p_j),\quad\mathrm{BISC}=\frac{1}{|\mathcal{P}_{\mathrm{bg}}|}\sum_{(i,j)\in\mathcal{P}_{\mathrm{bg}}}\!\!\!\!\cos(p_i,p_j),
\end{equation}
where $\cos(p_i,p_j)=\frac{p_i\cdot p_j}{\|p_i\|_2\|p_j\|_2}$, and $|\cdot|$ denotes the total number of predicted pairs in the respective set. Higher CISC and BISC scores indicate better preservation of identity and background consistency across complex multi-shot narratives.

\noindent\textbf{Shot Transition Accuracy (STA).}
To evaluate whether the generated video follows the prescribed shot-transition structure, we introduce Shot Transition Accuracy (STA), which jointly measures the correctness of the number of transitions and the temporal precision of their locations. 
Given the ground-truth transition frame set $\mathcal{G}=\{g_1,g_2,\dots,g_M\}$ and the predicted transition frame set $\mathcal{P}=\{p_1,p_2,\dots,p_N\}$ obtained by applying TransNetV2~\cite{TransNetv2} to the generated video, we first perform one-to-one matching between $\mathcal{G}$ and $\mathcal{P}$ to minimize the overall temporal deviation. For each matched pair $(p,g)$, the alignment error is measured by the absolute frame difference $|p-g|$. Unmatched ground-truth and predicted transitions are collectively treated as unaligned transitions, and are assigned a fixed penalty $\delta$. Let $\mathcal{N}$ denote the set of matched pairs, and let $N_{\mathrm{unmat}}$ denote the total number of unmatched transitions. We then define:
\begin{equation}
\mathrm{STA}=\exp\left(-\frac{\sum_{(p,g)\in\mathcal{N}}|p-g|+\delta\cdot N_{\mathrm{unmat}}}{N_{\mathrm{total}}}\right),
\end{equation}
where $N_{\mathrm{total}}$ is the total number of frames in the video. STA ranges in $(0,1]$, with a higher score indicating better agreement between the generated transition structure and the target timeline.

\subsection{Dataset Samples}
\begin{figure}[t]
\centerline{
\includegraphics[width=1.0\columnwidth]{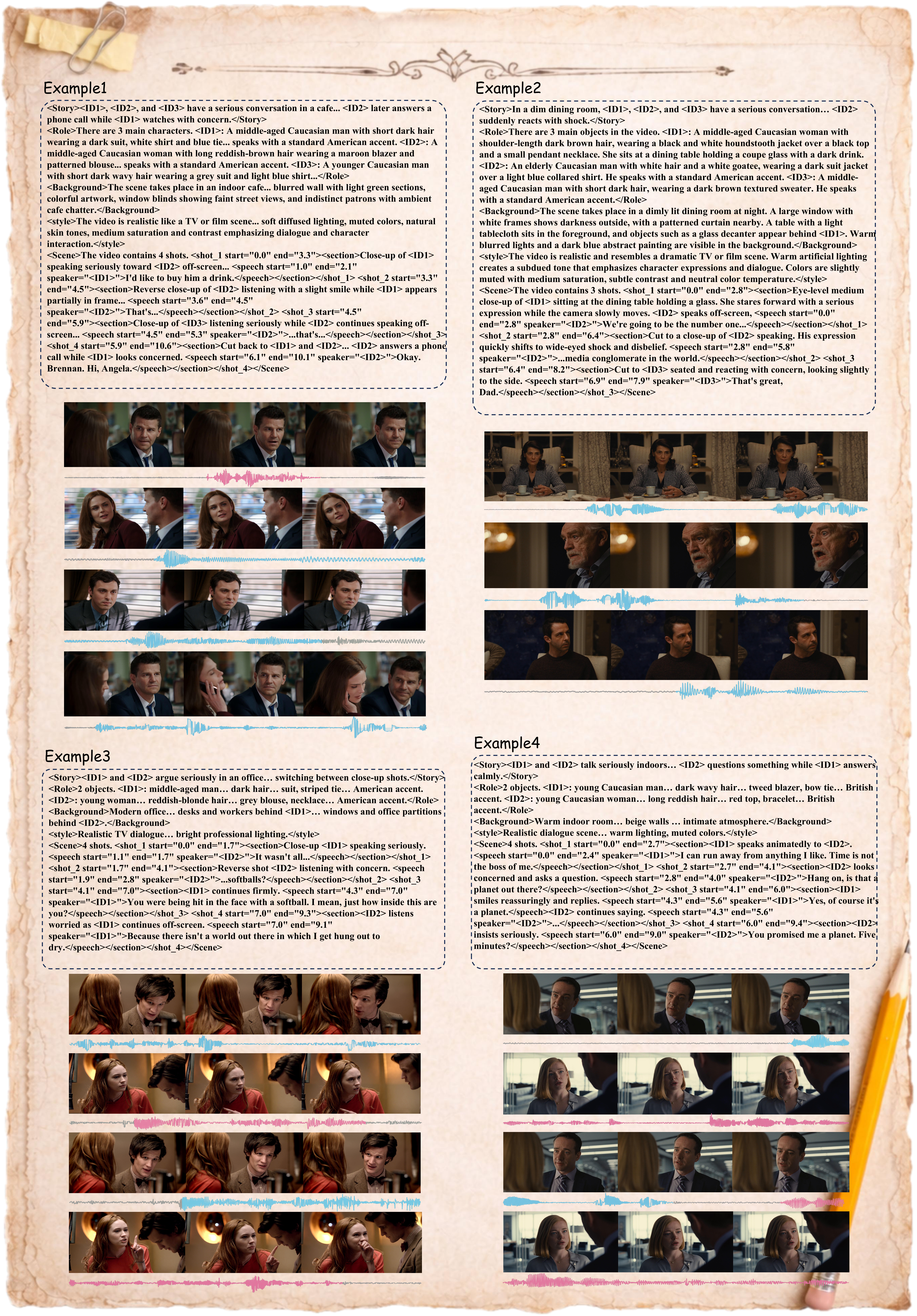}
}
\caption{
Data samples from our proposed MAVINSet, illustrating the alignment between hierarchical captions and multi-shot audio-visual sequences.
}

\label{fig:supp_example}
\end{figure}

As introduced in Sec.~3 of the main paper, we propose MAVINSet tailored for multi-shot audio-visual generation. 
To provide an intuitive understanding of the data structure, we randomly select samples and present them in \cref{fig:supp_example}, showcasing the detailed hierarchical captions paired with their corresponding audio-visual shots.

\clearpage
{
\small
\bibliographystyle{splncs04}
\bibliography{main}
}

\end{document}

\endgroup
\fi

\end{document}